\PassOptionsToPackage{table}{xcolor}
\documentclass[10pt]{article} 
\usepackage[preprint]{tmlr}

\usepackage{amsmath,amsfonts,bm}

\def\eqref#1{equation~\ref{#1}}

\def\1{\bm{1}}

\DeclareMathAlphabet{\mathsfit}{\encodingdefault}{\sfdefault}{m}{sl}
\SetMathAlphabet{\mathsfit}{bold}{\encodingdefault}{\sfdefault}{bx}{n}

\usepackage{float}
\usepackage{hyperref}
\usepackage{url}
\usepackage{graphicx}
\usepackage{subcaption}
\usepackage{booktabs}
\usepackage{multirow}
\usepackage{tabularx}      
\usepackage{array}   
\usepackage[table]{xcolor}
\usepackage{enumitem}
\usepackage{algorithm}
\usepackage{algorithmic}
\usepackage{nicematrix}
\newcommand{\err}[1]{\textcolor{red}{#1}}
\definecolor{addgreen}{RGB}{0,0,0}
\definecolor{colcopy}{RGB}{74,144,226}
\definecolor{colrecomb}{RGB}{80,227,194}
\definecolor{colgen}{RGB}{245,166,35}
\makeatletter
\long\def\NEW#1{{\color{addgreen}#1}}
\long\def\CHG#1{{\color{black}#1}}

\makeatother
\definecolor{highRelBlue}{RGB}{222, 235, 247}  
\definecolor{lowRelRed}{RGB}{255, 229, 229}    

\newcommand{\heatmap}[1]{%
    \ifdim #1pt>80pt \cellcolor{green!60}%
    \else\ifdim #1pt>60pt \cellcolor{green!30}%
    \else\ifdim #1pt>40pt \cellcolor{yellow!30}%
    \else\ifdim #1pt>20pt \cellcolor{orange!30}%
    \else \cellcolor{red!30}%
    \fi\fi\fi\fi #1%
}
\title{Large Language Models as Automatic Annotators and Annotation Adjudicators for Fine-Grained Opinion Analysis}

\author{\name Gaurav Negi \email gaurav.negi@insight-centre.org \\
      \addr Data Science Institute\\
      University of Galway
      \AND
      \name Al Waskow \email margaret.waskow@insight-centre.org \\
      \addr Data Science Institute\\
      University of Galway
      \AND
      \name John P. McCrae \email john.mccrae@insight-centre.org \\
      \addr Data Science Institute\\
      University of Galway
      \AND
      \name Omnia Zayed \email omnia.zayed@insight-centre.org \\
      \addr Data Science Institute\\
      University of Galway
      \AND
      \name Paul Buitelaar \email paul.buitelaar@insight-centre.org \\
      \addr Data Science Institute\\
      University of Galway
      }

\def\month{MM}  
\def\year{YYYY} 
\def\openreview{\url{https://openreview.net/forum?id=XXXX}} 

\begin{document}

\maketitle

\begin{abstract}
Fine-grained opinion analysis of text provides a detailed understanding of expressed sentiments and their targets. Although this level of detail is valuable, annotating opinions in datasets for model training requires considerable human effort and substantial cost, especially across diverse domains and real-world applications. To address this shortage of domain-specific labelled datasets, we explore the feasibility of LLMs as automatic annotators for fine-grained opinion analysis. We use a declarative annotation pipeline, an approach that reduces the variability of manual prompt engineering when using LLMs to identify fine-grained opinion spans in text. We also present a dedicated methodology for an LLM to adjudicate multiple labels and produce final annotations\NEW{, benchmarked against exact, flexible, and element-wise variants of a rule-based voting aggregator}. We trial the pipeline with models of different sizes for the Aspect Sentiment Triplet Extraction (ASTE) and Aspect-Category-Opinion-Sentiment (ACOS) analysis tasks. Our results reveal a critical performance bifurcation: LLMs are reliable at the span level yet struggle to reproduce the relational structures that connect those spans faithfully. This suggests that LLMs are better positioned as high-fidelity annotation assistants and data augmentation tools to expand fine-grained opinion-annotated datasets, rather than replacing human annotators entirely.
\end{abstract}

\section{Introduction}
Opinion analysis has evolved beyond simple sentiment polarity (positive, negative, neutral) to fine-grained formulations that also capture the contextual elements of an opinion. Consider the following example:
    
\textbf{Example}:\textit{``I had hoped for better battery life, as it had only about 2--1/2 hours doing heavy computations (8 threads using 100\% of the CPU).''}

\noindent In the example above, a \textit{negative} sentiment is expressed according to sentence- or document-level annotations. But with a fine-grained formulation, the following additional annotations are desired: (i) an \textbf{opinion span} expressing the polarity, in this example that is  \textit{``hoped for better''}, (ii) an explicit instantiation of the target in the text called an \textbf{aspect term}  (in the example: \textit{``battery life''}), They together form an aspect sentiment triplet as specified according to the Aspect Sentiment Triplet Extraction (ASTE) task \citep{grid_at_op_sentiment_2020}. Adding coarse-grained aspect categories (e.g., \textit{``Battery\#Operational\_performance''} in the example above) to identify the target entity and attribute enables the capture of implicit targets that lack explicit aspect terms, adhering to ACOS quadruple guidelines \citep{acos_extract_classify}.

Existing datasets for fine-grained opinion mining tasks are derived from SemEval tasks and are relatively small, domain-specific, and collected exclusively from review websites. As highlighted in \citep{absa_survey_2022}, these datasets are helpful in comparing predictive methods and systems, but are not sufficient for addressing real-world use cases. Subjective opinion annotation by human annotators is a non-trivial task and requires strict adherence to annotation guidelines, as documented in \citep{toprak_2010}. While human annotation remains the gold standard for opinion annotation, its lack of scalability motivates the assessment of automatic annotation approaches. 

In this work, we investigate the potential of LLMs to automatically annotate text with fine-grained opinions. The LLMs have been shown to be few-shot learners \citep{llm_zero_shot_2022} and to exhibit instruction-following and reasoning capabilities \citep{llm_reason_2024}. Through our study, we enquire into the extent to which they can successfully annotate fine-grained opinions \CHG{and their failure modes}. Understanding the efficacy of LLM-driven annotations can help us make an informed decision on the extent to which these models can fulfil the opinion annotator role or assist human annotators.


We also propose a dedicated LLM-based Adjudication approach that combines multiple opinion annotations for a given text input to resolve inter-annotator disagreement and obtain final opinion annotations. This is inspired by the process of annotation adjudication \citep{annotation_science_2010}, which includes making a final judgement after considering the conflicting annotations from multiple annotators. Commonly used adjudication methods, such as traditional majority voting, assume a discrete label set. However, ASTE and ACOS opinions form open-ended triplets and quadruples, \CHG{preventing direct application of traditional majority voting} \citep{annotation_difficulty_2021}. \NEW{We therefore adapt traditional voting into two rule-based set-level and element-wise aggregation baselines to compare against our LLM-based adjudication.} Our main contributions include:
\begin{itemize}[noitemsep, topsep=2pt, leftmargin=1.9em]
    \item \textbf{Declarative LLM annotation pipeline}: We propose a schema-constraint pipeline that uses LLMs declaratively, without hand-crafting prompts, thereby enabling a more systematic interaction for extracting opinions expressed with an ASTE and ACOS specification.

    \item \textbf{Annotation adjudication}: We introduce a dedicated method for aggregating annotations from multiple LLM-driven annotators to produce final annotations using an LLM as an annotation adjudicator.\NEW{ We benchmark this adjudicator against exact, flexible, and element-wise variants of a voting aggregator adapted to the combinatorial output space of ASTE and ACOS.}

    \item \textbf{Assessment of LLM at different scales}: We experiment with multiple LLMs with varying scales in our pipeline. This establishes an understanding of the trade-off between scale and desired efficiency in automatically annotating opinions.

    \item \textbf{Empirical analysis of LLM annotation reliability}: We evaluate opinion triplets and quadruples using element-wise and pairwise metrics, revealing that while LLMs reliably extract individual spans, their alignment with human annotations degrades markedly as relational complexity increases.
\end{itemize}
\section{Related Work}
\paragraph{Opinion Formulations} Opinion\footnote{Unless stated otherwise, we use the term opinion as a broad concept that covers sentiment and its associated information, such as opinion target and the person who holds the opinion, and use the term sentiment to mean only the underlying positive, negative, or neutral polarity implied by opinion.} Mining and sentiment analysis have been well explored in natural language processing. Aspect-Based Sentiment Analysis (ABSA), the foundation of fine-grained opinion mining, evolved from feature-based summarisation \citep{hu_mining_2004,DBLP:books/sp/mining2012/LiuZ12}, which involves extracting and summarising opinions about features (attributes/ keywords).  In this paper, we experiment with two fine-grained and extractive formulations of downstream ABSA tasks: Aspect Sentiment Triple Extraction (ASTE) \citep{aste_jet,grid_at_op_sentiment_2020} and Aspect-Category-Opinion-Sentiment Quadruple (ACOS) extraction\citep{acos_extract_classify}.

\paragraph{LLMs as Annotators} Key steps in dataset creation for building any intelligent system include data selection, guideline development, human annotation, quality estimation, and adjudication \citep{datainwild_2024}. With the improvement of the general ability of LLMs, they have been investigated for annotating various subjective tasks, including span annotations\citep{annot_spans_2025}, argument quality annotations\citep{argue_annot_2024} and propaganda span annotations\citep{annotate_propaganda_2024}. In this work, we extend this line of investigation to fine-grained opinion annotation, exploring LLMs as annotators using inference-time adaptation. Unlike supervised approaches that require task-specific fine-tuning, our method optimises prompts rather than model weights, operating in the same training-free spirit as earlier rule-based methods \citep{qiu_opinion_2011} while leveraging the broader reasoning capabilities of modern LLMs.

\paragraph{Prompt Engineering} LLMs are known to be sensitive to the prompts \citep{zhuo-etal-2024-prosa} and exhibit a higher level of variability in performing complex tasks \citep{llm_reproduce_2025}. To minimise the spurious interaction between the prompt and model selection as a confounding variable, we adopt \textbf{DSPy} \citep{dspy_2024} and the approach of \textit{``programming LLMs''} rather than prompting them naively. It shifts the focus from performing string manipulation on prompt strings to programming with structured, declarative modules that automatically prepare prompts using a small sample of annotated examples. \NEW{To aggregate these programmed outputs, we avoid single-model redundancy techniques such as self-consistency \citep{wang2023selfconsistency}. Instead, diverse annotators are combined. We evaluate fixed-rule voting aggregation and propose an LLM-as-adjudicator approach, inspired by stacked generalisation \citep{stacked_general_1992}, to synthesise distinct model annotations and resolve complex relational structures.}

\section{Dataset}
\label{sec:data_model}
\begin{figure}[h]
\hspace{20pt}
\begin{subfigure}[t]{0.35\textwidth}
\caption{ACOS \& ASTE Specifications}
\label{fig:subim1}
\vspace{0pt}
\includegraphics[width=\linewidth]{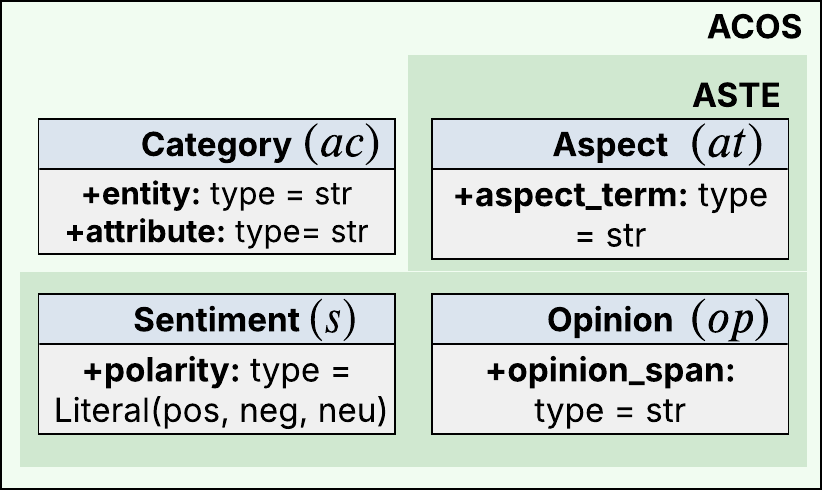}

\end{subfigure}
\hspace{30pt}
\begin{subfigure}[t]{0.45\textwidth}
\vspace{0pt}
\caption{Dataset Breakdown}
\label{fig:subim2}
\resizebox{\linewidth}{!}{
\begin{tabular}{ c c c c c }
\toprule
         Task & Dataset & Train & Dev & Test \\
    \hline
         \multirow{4}{*}{ASTE} & lap14 & 906 & 219 & 328\\
          & res14 & 1126 & 310 & 492 \\
          & res15 & 607 & 148 & 322\\
          & res16 & 857 & 210 & 326\\ \hline
        \multirow{2}{*}{ACOS} & laptop & 2934 & 326 & 816\\
          & restaurant & 1530 & 171 & 583\\
    \bottomrule
\end{tabular}
}

\end{subfigure}
\end{figure}
 \begin{table}[H]
\centering
\caption{Fine-Grained Opinion Examples}
\label{tab:aste_eg}

\resizebox{0.85\linewidth}{!}{%
\begin{tabular}{l l l}
\toprule
\textbf{Task} & \textbf{Input(s)} & \textbf{Output}\\
\midrule

\multirow{2}{*}{ACOS} & \textit{``I  have eaten here three times and have found the quality} & (\textcolor[HTML]{CC9900}{quality}, \textcolor[HTML]{38761D}{excellent}, \textcolor[HTML]{351C75}{Pos}, \textcolor[HTML]{351C75}{food\#quality}) \\

 & \textit{and variety of the fish to be excellent''} &  (\textcolor[HTML]{CC9900}{variety}, \textcolor[HTML]{38761D}{excellent}, \textcolor[HTML]{351C75}{Pos}, \textcolor[HTML]{351C75}{food\#style})\\
 
\midrule
\multirow{2}{*}{ASTE} & \multirow{2}{*}{\textit{``moules were excellent, lobster ravioli was very salty.''}}  & (\textcolor[HTML]{CC9900}{moules}, \textcolor[HTML]{38761D}{excellent}, \textcolor[HTML]{351C75}{Pos})\\ 
& & (\textcolor[HTML]{CC9900}{lobster ravioli}, \textcolor[HTML]{38761D}{salty}, \textcolor[HTML]{351C75}{Neg})\\\midrule
\end{tabular}
}
\end{table}

The datasets used in our experiments are described in Table \ref{fig:subim2}. Because fine-tuning LLMs is not an objective of our annotation pipeline, we do not utilise training splits. Instead, we use the development split to provide In-context Learning (ICL) examples, which are included in the optimised prompt. The test splits remain unchanged, allowing us to evaluate performance across multiple LLMs in our experiments. The schema in Figure \ref{fig:subim1} illustrates the relationship between the two formulations we explore in this work.

\begin{itemize}[noitemsep, topsep=2pt, leftmargin=1.75em]
    \item \textbf{ASTE}: Aspect Sentiment Triple Extraction (ASTE) falls under the umbrella of ABSA tasks. Proposed by \cite{peng_2020}, it is a task for extracting opinion triplets, thus providing one-shot answer to: \textbf{What} target is being discussed, \textbf{How} is the sentiment (e.g. positive) and \textbf{Why} is this sentiment (e.g. ``\textit{excellent}'').
    \item \noindent \textbf{ACOS}: This specification Aspect-Category-Opinion-Sentiment(ACOS) Quadruple Extraction, to extract aspect, aspect category, opinion and sentiment as quadruples in text and provide full support for aspect-based sentiment analysis with implicit aspects and opinions \footnote{Opinion in ACOS task refers to the span that expresses sentiment.}. The primary focus of this formulation is the fine-grained analysis of the opinion target. This task is more challenging than ASTE because it not only adds a category on top of the existing opinion triplet, but also takes into account implicit cases (i.e. missing explicit aspect terms or opinion spans) \citep{acos_extract_classify}.
\end{itemize}

\section{Methodology}
The fine-grained opinion annotation pipeline has multiple stages: (i) Preparing and evaluating LLM-specific prompts, (ii) Redundant Opinion annotation using multiple LLMs, and (iii) Annotation adjudication using LLM to create collaborative opinion annotations. \NEW{In addition, we implement rule-based voting aggregators (Section~\ref{sec:vote_adj}) that serve as baselines against which the LLM-based adjudication is compared.}

\subsection{Problem Definition}
On a conceptual level, the process of annotating opinions using the fine-grained opinion formulations involves identifying a set of opinions expressed in the input text $T_i$. As with human annotators, the process generates multiple annotations for the same input $T_i$, followed by adjudication that resolves disagreements between annotators to produce final annotations. These steps can be formalised as:
\smallskip

\noindent \textbf{LLM as opinion annotators.} Given the input text $T_i$ from corpora $\mathcal{T}$, our objective is to extract all expressed opinions $\mathcal{O}_i =\{o_1, o_2,.. o_{n}\}$. The elements of single opinion ($o_j$) is task dependant: 
\begin{itemize}[noitemsep, topsep=2pt, leftmargin=1.75em]
    \item ASTE opinions are formulated as the triple $o_{j} = (at,s, op)$, where $at$ is the aspect term, $s$ is the sentiment and $op$ is the opinion term.
    \item ACOS opinions are quadruples $o_j = (at,s, op, ac)$, where $at$, $s$ and $op$ encode similar concepts as ASTE and $ac$ is aspect category made up of entity $E$ and attribute $A$ (expressed as $E\#A$).
\end{itemize}
The LLM-driven annotation pipeline maps input corpora $\mathcal{T}$ to $\mathcal{O}^k$ using $k$ LLM-based annotators.

\paragraph{Opinion Adjudication.} We evaluate two main strategies for opinion adjudication:
\begin{itemize}[noitemsep, topsep=2pt, leftmargin=1.75em]
    \item \textbf{LLM-Based Adjudication ($\mathcal{A}\mathrm{dj}_{\mathrm{LLM}}$):} LLM-based adjudication to merge opinion sets $(\mathcal{O}^{1}, \dots, \mathcal{O}^{k})$ produced by $k$ annotators given the source text $\mathcal{T}$:
$\mathcal{A}\mathrm{dj}_{\mathrm{LLM}} : (\mathcal{T}, \mathcal{O}^{1}, \dots, \mathcal{O}^{k}) \rightarrow \mathcal{O}$. This formulation is analogous to stacking \citep{stacked_general_1992} in ensemble learning.
    \item \NEW{\textbf{Voting Adjudication Baselines (${\mathrm{Vot}}$):} Rule-based counterparts $\mathcal{A}\mathrm{dj}_{\mathrm{LLM}}$that operate strictly on the annotation sets:
    $\mathrm{Vot} : (\mathcal{O}^{1}, \dots, \mathcal{O}^{k}) \rightarrow \mathcal{O}$
    These retain only opinions supported by a minimum annotator threshold (Section~\ref{sec:vote_adj}) and are used for benchmarking purposes.}
\end{itemize}

\subsection{Declarative Annotation and Adjudication Pipeline}\label{sec:pipeline}
\begin{figure}[thbp]
\begin{center}
\includegraphics[width=0.8\linewidth]{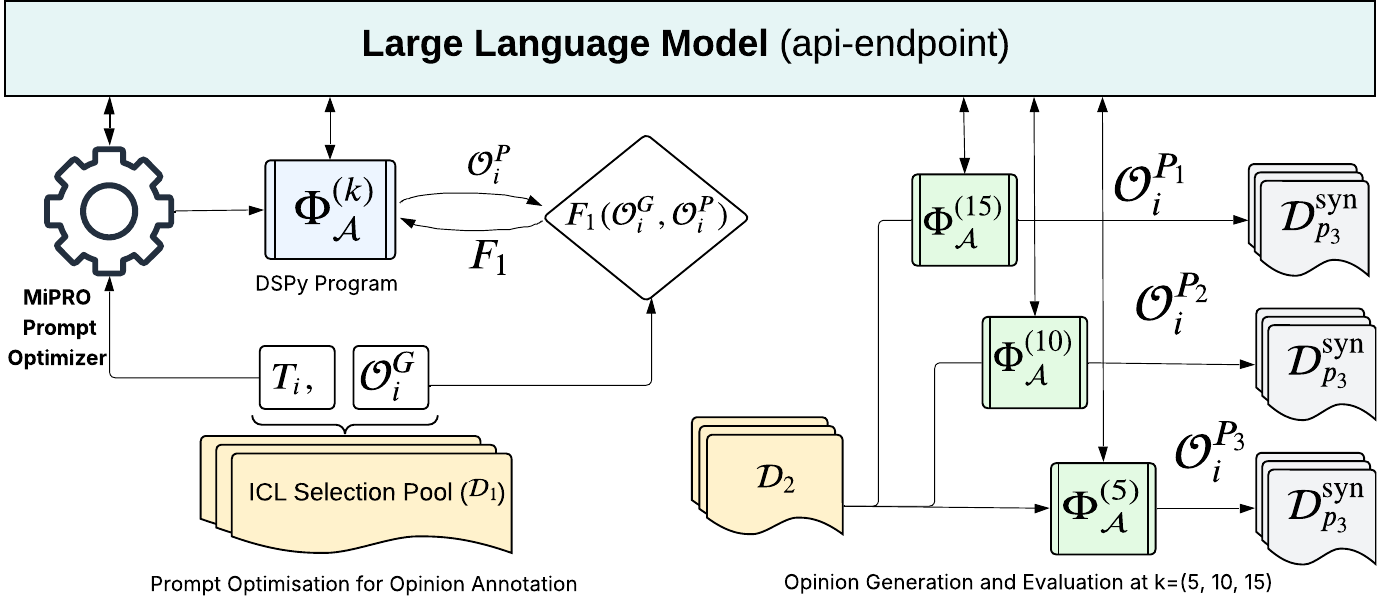}
\caption{Overview of the LLM-based annotation architecture using DSPy. 
\textbf{Left:} The synthesis stage, where the DSPy program $\Phi_{\mathcal{A}}$ utilizes $k$ human-annotated examples $\mathcal{O}_i^{G}$ from an ICL selection pool $\mathcal{D}_1$ to construct optimized prompts. 
\textbf{Right:} The annotation stage, where predicted opinions $\mathcal{O}^P_{i}$ are generated for input texts $T_i$ within the synthetic dataset $\mathcal{D}^{syn}$. 
The pipeline is evaluated across varying exemplar-counts $k \in \{5, 10, 15\}$ on a distinct development split $\mathcal{D}_2$.}
\label{fig:annot1}
\end{center}
\end{figure}

We structure the generation and adjudication steps into a multi-stage declarative pipeline powered by a DSPy program. Both the annotation and adjudication stages rely on the MIPROv2 prompt optimiser to systematically refine the instructions and in-context examples  \citep{mipro_2024}. The pipeline is trained and evaluated across two data splits: $D_1$ (the first half of the development set) and $D_2$ (the second half of the development set).

\paragraph{Stage 1: Optimising the Annotation Pipeline on $D_1$}

In the first stage, the objective is to optimise the LLM-based opinion-generation pipeline, denoted by $\Phi_{\mathcal{A}}^{(k)}$, to extract opinions from the input text $T_i$. We utilise MIPROv2 on $D_1$ to systematically optimise prompts across three few-shot levels, $k \in \{5, 10, 15\}$. The optimiser explores different combinations of instructions and in-context examples sampled from $D_1$ to maximise the quality of the extracted opinions.

\paragraph{Stage 2: Redundant Generation and Evaluation on $D_2$}

Once the generation programs are optimised, we apply them to the unseen texts in $D_2$ to fulfil two purposes. First, we validate the performance of the generated pipelines at different $k$ values by computing their $F_1$ scores against the ground-truth annotations, $F_1(\mathcal{O}_i^G, \mathcal{O}_i^p)$, thereby determining the optimal $k$ setting. \NEW{We emphasise that this validation on $D_2$ is an intermediate configuration step, analogous to hyper-
parameter search on a development set; all results reported in Sections~\ref{sec:results} and~\ref{sec:analysis} are computed exclusively on the held-out test splits.} 

Second, we use these optimised programs to perform redundant data generation over $D_2$. For a given input text $T_i$, this process yields multiple sets of predicted candidate annotations (denoted as $\mathcal{O}_i^{P_1}, \mathcal{O}_i^{P_2}, \mathcal{O}_i^{P_3}$). This results in a comprehensive dataset containing the original text, redundant LLM predictions, and gold annotations.

\paragraph{Stage 3: Optimising the Adjudication Pipeline}

\begin{figure}[thbp]
\begin{center}
\includegraphics[width=.7\linewidth]{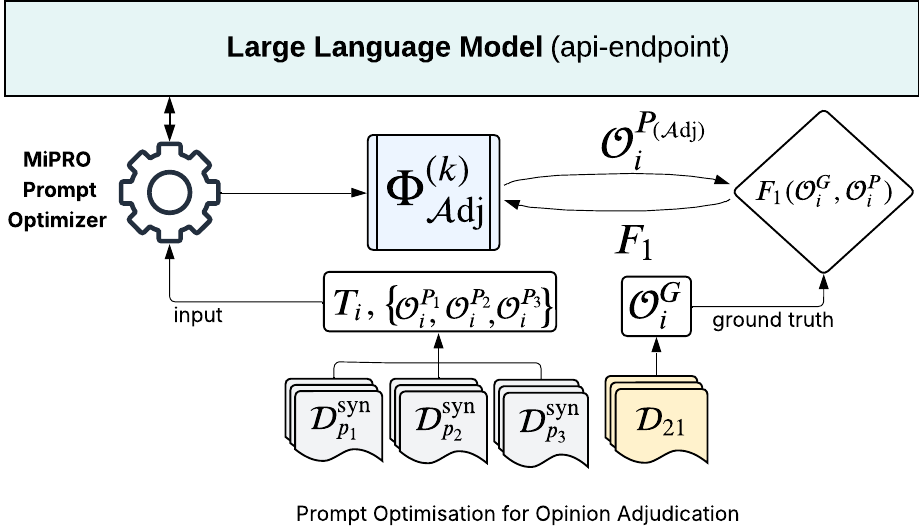}
\caption{LLM-based adjudication DSPy pipeline. Annotations at $k \in \{5, 10, 15\}$ are combined with the original split ($\mathcal{D}_2$) to generate the adjudication DSPy program ($\Phi_{\mathcal{A}dj}^{(k)}$)}.
\label{fig:adj1}
\vspace{-20pt}
\end{center}
\end{figure}

In the final stage, we train the adjudication pipeline ($\Phi_{\mathcal{A}dj}$). This phase exclusively utilises the data formulated in Stage 2 from the $D_2$ split. The inputs to the adjudication model include the original source text $T_i$ and the redundant predicted sets $\{\mathcal{O}_i^{P_1}, \mathcal{O}_i^{P_2}, \mathcal{O}_i^{P_3}\}$. 

Using MIPROv2, the adjudicator is optimised to evaluate and resolve conflicts among the redundant opinions to produce a final, synthesised output prediction, $\mathcal{O}_i^{P_{(\mathcal{A}dj)}}$. The optimisation is driven by maximising the $F_1$ score between the final adjudicated output and the ground truth $\mathcal{O}_i^G$ present in $D_2$. \NEW{The adjudicator performs a joint extraction conditioned on the input text and the redundant candidate sets, emitting the full adjudicated annotation set in a single pass rather than selecting among candidates element by element.} 

\NEW{
\subsubsection{Voting Adjudication}\label{sec:vote_adj}
We propose two rule-based voting aggregators as adjudication performance baselines:

\noindent \textbf{Set-Level Voting}: Illustrated in Algorithm~1a, this method votes over the global set of annotator--tuple pairs $X = \bigcup_{i=1}^M \{(i, e)\}$ across $M$ annotators using two matching variants: (i) \textbf{exact matching}, which groups tuples with identical normalised keys (including the aspect category for ACOS quadruples), and 
(ii) \textbf{flexible matching}, which handles span discrepancies via an undirected graph $G = (X, E)$. 
An edge connects $(i, e_a)$ and $(j, e_b)$ iff $s_a = s_b$ and both target and opinion spans overlap ($W(t_a) \cap W(t_b) \neq \emptyset \land W(o_a) \cap W(o_b) \neq \emptyset$). Implicit (null) spans map to a unique token to prevent matching explicit spans. Using Union-Find, we extract connected components as clusters $\mathcal{C}$ and discard those supported by fewer than $\theta$ unique voters ($|\{ i \mid (i, e) \in C \}| < \theta$). 
For each surviving cluster $C$, the majority surface form $e^*$ is selected as representative, breaking frequency ties by the longest explicit opinion span. 

\noindent \textbf{Element-Wise Voting}: Illustrated in Algorithm~1b, this method operates on individual tuple fields rather than whole tuples. Tuples are aligned greedy-style (strongest alignment first, max one tuple per annotator) if $\ge 50\%$ of fields agree and either an explicit span overlaps or the tuples are identical. 
Each aligned cluster emits a single tuple whose fields are independently set by majority vote, breaking ties by highest annotator rank then longest surface form. Unlike set-level voting, this variant can \textbf{recombine fields} from different annotators, though it never introduces unproposed elements.}

\begin{table}[htbp]
\centering
\scriptsize
\setlength{\tabcolsep}{0pt}
\begin{minipage}[t]{0.48\textwidth}
\centering
\textcolor{addgreen}{\textbf{(a) Consensus Voting}}\par\smallskip
\begin{tabular}{p{\linewidth}}
\hline
\textbf{In:} $\{A_1,\dots,A_M\}$, $\sim\!\in\!\{\text{exact},\text{flex}\}$, $\theta$ \\
\textbf{Out:} consensus set $Y$ \\
\hline
\begin{tabbing}
xx \= \; \= \; \= \kill
1. \> $X \leftarrow \bigcup_i \{(i,e): e\in A_i\}$ \\
2. \> \textbf{if} $\sim$ exact \textbf{then} \\
3. \> \> $\mathcal{C}\leftarrow$ partition $X$ by $\operatorname{norm}(e)$ \\
4. \> \textbf{else} \\
5. \> \> $\mathcal{C}\leftarrow$ Components on span overlap \\
6. \> $Y \leftarrow \emptyset$ \\
7. \> \textbf{for each} $C\in\mathcal{C}$ \textbf{do} \\
8. \> \> \textbf{if} $|\{i:(i,e)\in C\}|\ge\theta$ \textbf{then} \\
9. \> \> \> $e^*\!\leftarrow\!\operatorname{argmax}_u(\operatorname{freq}(u),\operatorname{len})$ \\
10. \> \> \> $Y \leftarrow Y \cup \{e^*\}$ \\
11. \> \textbf{return} $Y$
\end{tabbing} \\
\hline
\end{tabular}
\end{minipage}\hfill
\begin{minipage}[t]{0.48\textwidth}
\centering
\textcolor{addgreen}{\textbf{(b) Element-wise Voting}}\par\smallskip
\begin{tabular}{p{\linewidth}}
\hline
\textbf{In:} $\{A_1,\dots,A_M\}$ over fields $F$; $\theta$ \\
\textbf{Out:} consensus set $Y$ \\
\hline
\begin{tabbing}
xx \= \; \= \; \= \kill
1. \> $X \leftarrow \bigcup_i \{(i,e): e\in A_i\}$ \\
2. \> $E\leftarrow\{(a,b): \operatorname{ann}(a)\ne\operatorname{ann}(b),$ \\
3. \> \> $\textsc{Agree}(e_a,e_b)\}$ \\
4. \> $\mathcal{C}\leftarrow$ Components$(X,E)$, greedy, \\
5. \> \> $\le 1$ tuple/annotator \\
6. \> $Y \leftarrow \emptyset$ \\
7. \> \textbf{for each} $C\in\mathcal{C}$ \textbf{do} \\
8. \> \> \textbf{if} $|\{i:(i,e)\in C\}|\ge\theta$ \textbf{then} \\
9. \> \> \> $e^*[f]\leftarrow\textsc{Maj}_f(C),\ \forall f\!\in\!F$ \\
10. \> \> \> $Y \leftarrow Y \cup \{e^*\}$ \\
11. \> \textbf{return} $Y$
\end{tabbing} \\
\hline
\end{tabular}
\end{minipage}
\caption*{\NEW{\textbf{Algorithm 1:}  \textbf{(a)} Set-level voting groups full tuples via exact or flexible matching. \textbf{(b)} Element-wise voting aligns tuples across shared fields and constructs representative tuples field-by-field.}}
\label{alg:vote}
\end{table}

\section{Experimental Setup}\label{sec:setup}
\paragraph{Model Configuration} In our experiments, we used models of three different sizes: mini (~4B parameters), small (~14B parameters), and medium (~32B parameters). The assignment of LLMs as $\mathcal{A}_1, \mathcal{A}_2, \mathcal{A}_3$ is carried out based on the model's performance on evaluation samples, and for all cases, the adjudication is performed by $\mathcal{A}_1$, which is the notation assigned to the best-performing LLM-based annotator.

\begin{table}[htbp]
    \centering
    \footnotesize 
    \caption{LLMs used as annotators. \textsuperscript{\dag} Used with both Small and Med}
    \label{tab:tdatasets}
    \resizebox{\linewidth}{!}{
    \begin{tabular}{c | c | c | c}
        \toprule
        \textbf{Annotator} & \textbf{Mini ($\sim$4B)} & \textbf{Small ($\sim$14B)} & \textbf{Med ($\sim$ 30B)} \\
        \midrule
        $\mathcal{A}_1$ & Qwen3-4B-Thinking & Qwen3-14B & Qwen/Qwen3-30B-A3B-Thinking-2507-FP8 \\
        $\mathcal{A}_2$ & MiniCPM3-4B & openai/gpt-oss-20b \textsuperscript{\dag}  & unsloth/DeepSeek-R1-Distill-Qwen-32B-bnb-4bit\\
        $\mathcal{A}_3$ & Phi-4-mini-reasoning & DeepSeek-R1-Distill-Qwen-14B& openai/gpt-oss-20b \textsuperscript{\dag} \\
        
        \bottomrule
    \end{tabular}
    }
    
\end{table}

We evaluate the pipeline across three model scales: \textbf{Mini}, \textbf{Small}, and \textbf{Med}(Medium). Within each scale, we consider three distinct LLMs (see Table \ref{tab:tdatasets}). For each model, we first identify its optimal configuration by validating performance at $k \in \{5, 10, 15\}$ on the $D_2$ split. The models are then ranked as $\mathcal{A}_1, \mathcal{A}_2, \text{ and } \mathcal{A}_3$ based on these $F_1$ validation scores. \footnote{dspy artefacts: \url{https://github.com/ANON-1331/anon-repo-annotation-opinion}}

The LLM  assigned to $\mathcal{A}_1$ is also selected for the adjudication stage because of its demonstrated efficacy on the annotation task. The adjudication process is analogous to appointing a high-performing human annotator as adjudicator \citep{annotation_science_2010}. To achieve this, we train a separate DSPy program and then optimise via MIPROv2, obtaining $\mathcal{A}\text{dj}$. This two-stage approach ensures that, while the redundant annotations are generated by a diverse ensemble, the final synthesis is performed by the most capable model in the tier, using a prompt specifically optimised to resolve inter-model disagreements. \NEW{However, while distinct pipelines with different DSPy programs are utilised, this selection criterion for adjudication might introduce cases of self-selection bias, mirroring the inherent challenges of the traditional annotation-adjudication design that motivated our architecture.}

\paragraph{Hardware Usage} Each experiment was run on a single Nvidia A40 GPU, with vLLM \citep{kwon_vllm_2023}  used to serve each model through the various steps of our pipeline, as required by DSPy for OpenAI-compliant API access. We keep the LLM hyperparameters constant across all models. The model's context window determines the input sequence length, and the generated sequence's output length is set to 16384. The pipeline performs inference-time adaptation at temperature 0\NEW{, individual runs are near-deterministic; we therefore rely on consistency across model scales, tasks, and datasets, together with the paired bootstrap tests of Appendix~\ref{sec:significance}.}

\paragraph{\textcolor{addgreen}{Cost and Throughput}} \NEW{The pipeline annotation process incurs costs of \$0.017 to \$0.036 per sentence, or \$5--\$28 per SemEval-scale split of 300--800 sentences. In comparison, \cite{epub79455} report that crowd-sourced ABSA costs \pounds0.41 ($\approx$\,\$0.55) per sentence for a single human pass, indicating that the pipeline achieves 15 to 30 times lower cost while providing three-fold redundancy. Operating on a single NVIDIA A40 at on-demand rates (\$0.29--\$0.44/h), full splits are triple-annotated and adjudicated in 18--65 unattended GPU-hours. According to reconstructed timestamps (Appendix~\ref{sec:throughput}) for the models in Table~\ref{tab:tdatasets}, annotator passes require 64 to 78 seconds, and adjudications require 25 to 59 seconds. Annotator timing data are available exclusively for the ACOS pass, which involves longer reasoning traces and a larger output space. Therefore, these figures are direct for ACOS and represent a conservative upper bound for ASTE.}

\paragraph{Data Usage} \CHG{As detailed in Section~\ref{sec:pipeline}, pipeline preparation strictly uses the development set: $\mathcal{D}_1$ for prompt optimization, and $\mathcal{D}_2$ for intermediate selection ($k$ selection, annotator ranking) and adjudicator optimization. LLMs are not fine-tuned, and training splits remain unused. All evaluations (Sections~\ref{sec:results} and~\ref{sec:analysis}) are computed on held-out SemEval test splits, preventing data leakage.}

\paragraph{Evaluation} We evaluate pipeline performance on the ASTE and ACOS tasks across three model scales (Mini, Small, and Medium). Because exact-match metrics are notoriously stringent for combinatorial tasks requiring joint span, category, and sentiment alignment \citep{barnesIfYouveGot2021}, we supplement standard evaluation with granular diagnostics. Specifically, we measure reference-based performance using exact-match, element-wise, and pairwise $F_1$ scores against ground-truth labels to isolate boundary versus relational errors. In tandem, we measure annotator consensus via Inter-Annotator Agreement (IAA) on span co-extractions to evaluate agreement among LLMs before adjudication.

\paragraph{Baseline Method} We adopt three foundational SFT paradigms as baselines: sequence tagging, grid-based span labelling, and encoder-decoder generation. These paradigms serve as the primary architectural foundations for supervised ASTE and ACOS methods. Within the proposed DSPy-based framework, foundational SFT models provide the most relevant point of comparison, enabling contextualization of emergent large language model (LLM) capabilities relative to task-specific supervision. \NEW{We compare against two external baselines: UnifiedABSA \citep{wangUnifiedABSAAnnotationDecoupled2024}, a supervised multi-task instruction-tuned model trained on annotations across all ABSA tasks (highly contrasting with our training-free paradigm), and the classical rule-based Double-Propagation method \citep{qiu_opinion_2011} for ACOS. Additionally, the voting aggregators in Section~\ref{sec:vote_adj} serve as rule-based adjudication baselines: $\mathcal{A}_{\mathrm{dj}}(Vot_{=})$, $\mathcal{A}_{\mathrm{dj}}(Vot_{\sim})$, and $\mathcal{A}_{\mathrm{dj}}(Vot_{\mathrm{el}})$ for exact, flexible, and element-wise variants ($\theta=2/3$), which are compared against the LLM-based adjudicator $\mathcal{A}_{\mathrm{dj}}(\mathcal{A}_1)$.}

\section{Results}\label{sec:results}
Triplet (ASTE) and quadruplet (ACOS) $F_1$ scores are separately reported against human gold references on held-out SemEval test splits. \NEW{To evaluate aggregation gains, paired bootstrap tests are conducted (Appendix~\ref{sec:significance}): $^{\ast}$ and $^{+}$ indicate significant improvements over all three- or two-tier annotators, respectively ($p<0.05$).}
\vspace{-5pt}
\begin{table}[h]
\centering
\caption{LLM-based annotator performance on ASTE (P/R/F1) using Inference-time adaptation methods. The baselines are SFT approaches using: sequence tagging method (BERT-JET by \cite{aste_jet}), grid tagging method (BERT-GTS by \cite{grid_at_op_sentiment_2020} and sequence-to-sequence generative method by \cite{yanUnifiedGenerativeFramework2021}.}
\label{tab:llm_aste_trip}
\resizebox{0.75\linewidth}{!}{
\setlength{\tabcolsep}{1.5pt}
\renewcommand{\arraystretch}{1.25}
\begin{tabular}{c|llcccc}
\toprule
& \multicolumn{2}{c}{\multirow{2}{*}{\textbf{Model}}}  & \textbf{Lap14} & \textbf{Res14} & \textbf{Res15} & \textbf{Res16} \\
\cmidrule(lr){4-4}  \cmidrule(lr){5-5} \cmidrule(lr){6-6} \cmidrule(lr){7-7}
& & & P / R / F1 & P / R / F1 & P / R / F1 & P / R / F1 \\
\toprule
\multirow{4}{*}{\rotatebox{90}{\textbf{SFT}}}
   & \multicolumn{2}{c}{JET}  & 55.39 / 47.33 / 51.04 &  70.56 / 55.94 / 62.40 & 64.45 / 51.96 / 57.53 & 70.42 / 58.37 / 63.83 \\
   & \multicolumn{2}{c}{GTS}  & 57.52 / 51.92 / 54.58 & 70.92 / 69.49 / 70.20 & 59.29 / 58.07 / 58.67 & 68.58 / 66.60 / 67.58 \\
   & \multicolumn{2}{c}{BART$_{ABSA}$}  & 61.41 / 56.19 / 58.69 & 65.52 / 64.99 / 65.25 & 59.14 / 59.38 / 59.26 & 66.60 / 68.68 / 67.62 \\
   & \multicolumn{2}{c}{\cellcolor{green!25}UnifiedABSA}  & - / - / \textbf{62.45} & - / - / \textbf{73.22} & - / - / 63.25 & - / - / \textbf{72.10} \\
\midrule
\multirow{21}{*}{\rotatebox{90}{\textbf{Inference-Time Adaptation}}} 
& \multirow{+7}{*}{Mini} 
& $\mathcal{A}_{1}$ & 42.34 / 43.39 / 43.15 & 62.27 / 64.98 / 63.86 & 50.96 / 60.60 / 55.11 & 62.43 / 70.81 / 66.36 \\
& & $\mathcal{A}_{2}$ & 34.05 / 38.08 / 35.95 & 54.71 / 60.16 / 57.31 & 41.70 / 55.46 / 47.61 & 52.42 / 67.51 / 59.01 \\
& & $\mathcal{A}_{3}$ & 21.63 / 25.92 / 23.59 & 36.18 / 41.22 / 38.53 & 30.12 / 40.87 / 34.68 & 40.65 / 51.27 / 45.34 \\
\cmidrule{3-7}
& & \cellcolor{green!25}$\mathcal{A}_{\mathrm{dj}}(Vot_{\approx})$ & 44.88 / 35.94 / 39.91\NEW{$^{+}$} & 64.99 / 57.98 / 61.29\NEW{$^{+}$} & 52.37 / 52.69 / 52.53\NEW{$^{+}$} & 65.72 / 68.03 / 66.85\NEW{$^{+}$} \\
& & \cellcolor{green!25}$\mathcal{A}_{\mathrm{dj}}(Vot_{=})$ & 57.09 / 35.19 / 43.54\NEW{$^{+}$} & 73.11 / 58.08 / 64.73\NEW{$^{+}$} & 61.63 / 51.65 / 56.21\NEW{$^{+}$} & 75.00 / 67.25 / \textit{70.91}\NEW{$^{\ast}$} \\
& & \cellcolor{green!25}$\mathcal{A}_{\mathrm{dj}}(Vot_{\mathrm{el}})$ & 40.32 / 41.90 / 41.10\NEW{$^{+}$} & 62.48 / 65.21 / 63.81$^{+}$ & 48.53 / 58.09 / 52.88\NEW{$^{+}$} & 62.50 / 72.12 / 66.97$^{+}$ \\
& & $\mathcal{A}_{\mathrm{dj}}(\mathcal{A}_{1})$ & 45.95 / 45.43 / 45.69\NEW{$^{\ast}$} & 67.44 / 67.85 / 67.64\NEW{$^{\ast}$} & 54.31 / 62.65 / 58.18\NEW{$^{\ast}$} & 64.28 / 71.92 / 67.89\NEW{$^{+}$} \\
\cmidrule{2-7}
& \multirow{7}{*}{Small} 
& $\mathcal{A}_{1}$ & 44.05 / 45.85 / 44.93 & 62.39 / 64.49 / 63.42 & 54.92 / 66.80 / 60.28 & 60.33 / 70.43 / 64.99 \\
& & $\mathcal{A}_{2}$ & 40.07 / 42.14 / 41.08 & 57.00 / 63.48 / 60.06 & 47.49 / 62.47 / 53.96 & 53.20 / 66.15 / 58.98 \\
& & $\mathcal{A}_{3}$ & 31.79 / 34.20 / 32.95 & 55.26 / 59.76 / 57.42 & 45.88 / 56.29 / 50.56 & 57.66 / 68.09 / 62.44 \\
\cmidrule{3-7}
& & \cellcolor{green!25}$\mathcal{A}_{\mathrm{dj}}(Vot_{\approx})$ & 45.01 / 40.85 / 42.82 & 63.90 / 63.77 / 63.84\NEW{$^{+}$} & 52.72 / 62.11 / 57.03\NEW{$^{+}$} & 61.68 / 69.84 / 65.51 \\
& & \cellcolor{green!25}$\mathcal{A}_{\mathrm{dj}}(Vot_{=})$ & 50.45 / 40.66 / 45.03\NEW{$^{+}$} & 68.80 / 63.46 / 66.02\NEW{$^{\ast}$} & 58.07 / 61.07 / 59.53\NEW{$^{+}$} & 61.51 / 68.09 / 64.63\NEW{$^{+}$} \\
& & \cellcolor{green!25}$\mathcal{A}_{\mathrm{dj}}(Vot_{\mathrm{el}})$ & 41.09 / 43.07 / 42.06 & 61.98 / 65.71 / 63.79\NEW{$^{+}$} & 51.74 / 64.60 / 57.46\NEW{$^{+}$} & 59.13 / 71.21 / 64.61 \\
& & $\mathcal{A}_{\mathrm{dj}}(\mathcal{A}_1)$ & 38.89 / 41.77 / 40.24 & 59.13 / 64.38 / 61.65 & 50.66 / 63.56 / 56.38 & 57.01 / 69.64 / 62.69 \\
\cmidrule{2-7}
& \multirow{7}{*}{Med} 
& $\mathcal{A}_{1}$ & 43.93 / 47.50 / 45.64 & 66.38 / 71.12 / 68.67 & 56.77 / 68.67 / 61.98 & 59.87 / 71.98 / 65.37 \\
& & $\mathcal{A}_{2}$ & 42.31 / 44.73 / 43.49 & 60.38 / 64.69 / 62.46 & 55.03 / 66.60 / 60.26 & 58.25 / 68.68 / 63.04 \\
& & $\mathcal{A}_{3}$ & 40.07 / 42.14 / 41.08 & 57.00 / 63.48 / 60.06 & 47.49 / 62.47 / 53.96 & 53.20 / 66.15 / 58.98 \\
\cmidrule{3-7}

& & \cellcolor{green!25}$\mathcal{A}_{\mathrm{dj}}(Vot_{\approx})$ & 45.13 / 44.54 / 44.83 & 65.91 / 67.50 / 66.69 \NEW{$^{+}$} & 56.76 / 66.59 / 61.29 & 59.62 / 68.67 / 63.83 \\
& & \cellcolor{green!25}$\mathcal{A}_{\mathrm{dj}}(Vot_{=})$ & 51.49 / 44.54 / \textit{47.77}\NEW{$^{+}$} & 70.32 / 67.70 / 68.99\NEW{$^{+}$} & 61.11 / 67.21 / \textbf{64.04}\NEW{$^{+}$}& 61.51 / 68.09 / 64.63 \\
& & \cellcolor{green!25}$\mathcal{A}_{\mathrm{dj}}(Vot_{\mathrm{el}})$ & 44.62 / 48.24 / 46.36 & 64.03 / 69.32 / 66.57\NEW{$^{+}$} & 55.28 / 69.07 / 61.41 & 57.10 / 68.87 / 62.43 \\
& & $\mathcal{A}_{\mathrm{dj}}(\mathcal{A}_1)$ & 45.14 / 47.50 / 46.81 & 68.81 / 70.42 / \textit{69.27}\NEW{$^{+}$} & 57.66 / 68.24 / 62.51 & 62.33 / 72.76 / 67.14\NEW{$^{+}$} \\
\bottomrule
\end{tabular}
}
\vspace{-8pt}
\end{table}
\subsection{ASTE}

\textbf{Overview.} LLM-based annotator performance displays significant domain asymmetry. \CHG{On Lap14, supervised models vastly outperform inference-time adaptation (UnifiedABSA at 62.45\% and $\text{BART}_{\text{ABSA}}$ at 58.69\% $F_1$, vs.\ 47.77\% for Medium $\mathcal{A}_{\mathrm{dj}}(Vot_{=})$). Across the restaurant datasets, this gap closes considerably: on Res15, Medium $\mathcal{A}_{\mathrm{dj}}(Vot_{=})$ (64.04\%) outperforms fine-tuned UnifiedABSA (63.25\%) without parameter updates, while Res16 exhibits a small margin (70.91\% Mini $\mathcal{A}_{\mathrm{dj}}(Vot_{=})$ vs.\ 72.10\%) and Res14 a moderate one (69.27\% Medium $\mathcal{A}_{\mathrm{dj}}(\mathcal{A}_1)$ vs.\ 73.22\%). Each domain-best result yields statistically significant improvements over individual annotators ($p < 0.05$, marked $^{\ast}$/$^{+}$), highlighting strong zero-shot generalisation.}

\textbf{LLM-As-ASTE Annotators} Although peak performance occurs in the \textbf{Med} tier, scaling between \textbf{Mini} and \textbf{Small} is non-monotonic (trend anomaly discussed below). Crucially, adjudication consistently outperforms the best individual model ($\mathcal{A}_1$). Employing $\mathcal{A}_{\mathrm{dj}}(\mathcal{A}_1)$ provides absolute $F_1$ gains over $\mathcal{A}_1$ across the four datasets of $+2.54/3.78/3.07/1.53\%$ for Mini models, and $+1.17/0.60/0.53/1.77\%$ for Medium models.

\NEW{\textbf{LLM vs. Voting Adjudication.} Voting and LLM adjudication exhibit clear precision-vs-recall trade-offs. Exact matching ($Vot_{=}$) acts as a strict precision filter (e.g., $75.00\%$ precision on Mini Res16; $57.09\%$ vs.\ $42.34\%$ $\mathcal{A}_1$ on Lap14) by discarding non-verbatim tuples, while flexible ($Vot_{\approx}$) and element-wise ($Vot_{\mathrm{el}}$) variants recover recall (e.g., $48.24\%$ on Med Lap14). Relative performance scales with model capacity:
\begin{itemize}[noitemsep, topsep=2pt, leftmargin=1.5em]
    \item \textbf{Mini Tier:} $\mathcal{A}_{\mathrm{dj}}(\mathcal{A}_1)$ leads on Lap14 ($45.69\%$), Res14 ($67.64\%$), and Res15 ($58.18\%$), while $Vot_{=}$ peaks on Res16 ($70.91\%$).
    \item \textbf{Small Tier:} LLM adjudication degrades below individual annotators ($40.24\%$ $F_1$ on Lap14), making $Vot_{=}$ the top aggregator ($66.02\%$ on Res14).
    \item \textbf{Medium Tier:} High precision gives $Vot_{=}$ peak scores on Lap14 ($47.77\%$) and Res15 ($64.04\%$), whereas higher recall ($70.42\%$ and $72.76\%$) pushes $\mathcal{A}_{\mathrm{dj}}(\mathcal{A}_1)$ ahead on Res14 ($69.27\%$) and Res16 ($67.14\%$).
\end{itemize}}

\textbf{Trend Anomaly} As previously discussed, LLM adjudication performance unexpectedly declines in the \textbf{Small} tier (using Qwen3-14B). Since exact voting ($Vot_{=}$) continues to outperform $\mathcal{A}_1$ on these annotations (e.g., $66.02\%$ vs.\ $63.42\%$ on Res14), the issue is attributable solely to the adjudicator LLM. A potential explanation is that long-chain reasoning ("thinking") states might outweigh parameter scale for this task \citep{dseek_2025}, as these states are utilised by the Mini and Medium models but absent in the Small 14B model. \NEW{Table~\ref{tab:small_adj_ablation} supports this hypothesis: replacing the 14B model with a smaller, reasoning-distilled adjudicator (Qwen3-4B-Thinking) on the same annotations improves performance and tends towards restoring performance gains across all datasets.}

\begin{table}[!ht]
\centering
\scriptsize
\caption{\NEW{Small-tier adjudicator ablation on ASTE (P / R / $F_1$).}}
\label{tab:small_adj_ablation}
\resizebox{0.85\linewidth}{!}{
\begin{tabular}{lcccc}
\toprule
\textbf{Adjudicator} & \textbf{Lap14} & \textbf{Res14} & \textbf{Res15} & \textbf{Res16} \\
\midrule
Qwen3-14B & 38.89 / 41.77 / 40.24 & 59.13 / 64.38 / 61.65 & 50.66 / 63.56 / 56.38 & 57.01 / 69.64 / 62.69 \\
Qwen3-4B-Thinking & 43.36 / 43.20 / \textbf{43.28} & 65.66 / 67.28 / \textbf{66.46} & 55.55 / 65.08 / \textbf{59.94} & 61.48 / 72.37 / \textbf{66.48} \\
\bottomrule
\end{tabular}}
\end{table}

\subsection{ACOS}
\begin{table}[ht!]
\centering
\caption{LLM-based annotator performance on ACOS (P/R/F1) using an inference-time adaptation method. The baselines include a rule-based method \citep{qiu_opinion_2011}, sequence-labelling methods JET-ACOS and Extract-classify\citep{acos_extract_classify} and sequence-to-sequence generation method (BART-CRN by \citep{xiongBARTbasedContrastiveRetrospective2023c}).}
\label{tab:llm_acos}
\resizebox{0.55\linewidth}{!}{
\setlength{\tabcolsep}{6pt}
\renewcommand{\arraystretch}{1}

\begin{tabular}{c|llcc}

\toprule
& \multicolumn{2}{l}{\textbf{Model}}  & \textbf{Laptop} & \textbf{Restaurant} \\
\cmidrule(lr){4-4} \cmidrule(lr){5-5}
& & & P / R / F1 & P / R / F1 \\
\toprule
\multirow{4}{*}{\rotatebox{90}{\textbf{SFT}}}
   & \multicolumn{2}{l}{JET-ACOS} & 44.52 / 16.25 / 23.81 & 59.81 / 28.94 / 39.01 \\
   & \multicolumn{2}{l}{Extract-classify} & 45.56 / 29.48 / 35.80 & 38.54 / 52.96 / 44.61 \\
   & \multicolumn{2}{l}{BART-CRN} & 48.16 / 31.83 / 38.32 & 50.84 / 47.10 / 48.90 \\
   & \multicolumn{2}{l}{UnifiedABSA} & - / - / \textbf{42.58} & - / - / \textbf{60.60} \\
\midrule
\multirow{22}{*}{\rotatebox{90}{\textbf{Inference-Time Adaptation}}} 
& \multicolumn{2}{l}{Double-Propagation}  & 13.04 / 05.70 / 08.00 & 34.67 / 15.08 / 21.04\\
\cmidrule{2-5}
& \multirow{7}{*}{Mini} 
& $\mathcal{A}_{1}$ & 13.18 / 12.97 / 13.07 & 32.11 / 28.71 / 30.31 \\
& & $\mathcal{A}_{2}$ & 11.65 / 12.63 / 12.12 & 21.27 / 21.51 / 21.39 \\
& & $\mathcal{A}_{3}$ & 01.36 / 01.90 / 01.58 & 04.62 / 05.89 / 05.18 \\
\cmidrule{3-5}
& & \cellcolor{green!25}$\mathcal{A}_{\mathrm{dj}}(Vot_{\approx})$ & 10.94 / 7.96 / 9.21 & 33.22 / 21.62 / 26.19\NEW{$^{+}$} \\
& & \cellcolor{green!25}$\mathcal{A}_{\mathrm{dj}}(Vot_{=})$ & 32.57 / 6.14 / 10.33 & 59.36 / 16.27 / 25.54\NEW{$^{+}$} \\
& & \cellcolor{green!25}$\mathcal{A}_{\mathrm{dj}}(Vot_{\mathrm{el}})$ & 13.21 / 13.15 / 13.18 & 30.59 / 27.18 / 28.79\NEW{$^{+}$} \\
& & $\mathcal{A}_{\mathrm{dj}}(\mathcal{A}_1)$ & 16.71 / 16.17 / 16.43\NEW{$^{\ast}$} & 32.90 / 30.45 / 31.63\NEW{$^{+}$} \\
\cmidrule{2-5}
& \multirow{7}{*}{Small} 
& $\mathcal{A}_{1}$ & 09.78 / 09.60 / 09.69 & 22.67 / 20.19 / 21.36 \\
& & $\mathcal{A}_{2}$ & 07.62 / 07.69 / 07.66 & 24.27 / 21.72 / 22.92 \\
& & $\mathcal{A}_{3}$ & 06.92 / 06.49 / 06.70 & 16.88 / 14.74 / 15.73 \\
\cmidrule{3-5}
& & \cellcolor{green!25}$\mathcal{A}_{\mathrm{dj}}(Vot_{\approx})$ & 9.26 / 8.22 / 9.00 & 28.83 / 20.74 / 24.13\NEW{$^{+}$} \\
& & \cellcolor{green!25}$\mathcal{A}_{\mathrm{dj}}(Vot_{=})$ & 16.94 / 5.36 / 8.15 & 42.67 / 17.79 / 25.12\NEW{$^{+}$} \\
& & \cellcolor{green!25}$\mathcal{A}_{\mathrm{dj}}(Vot_{\mathrm{el}})$ & 08.91 / 08.65 / 08.78 & 25.74 / 22.71 / 24.13\NEW{$^{+}$} \\
& & $\mathcal{A}_{\mathrm{dj}}(\mathcal{A}_1)$ & 10.94 / 11.07 / 11.01\NEW{$^{\ast}$} & 30.86 / 29.14 / 29.98\NEW{$^{\ast}$} \\
\cmidrule{2-5}
& \multirow{7}{*}{Med} 
& $\mathcal{A}_{1}$ & 15.77 / 15.74 / 15.75 & 29.37 / 27.51 / 28.41 \\
& & $\mathcal{A}_{2}$ & 09.63 / 09.94 / 09.78 & 17.48 / 17.46 / 17.47 \\
& & $\mathcal{A}_{3}$ & 07.62 / 07.69 / 07.66 & 24.27 / 21.72 / 22.92 \\
\cmidrule{3-5}
& & \cellcolor{green!25}$\mathcal{A}_{\mathrm{dj}}(Vot_{\approx})$ & 12.94 / 11.07 / 11.93\NEW{$^{+}$} & 26.66 / 21.51 / 23.81 \\
& & \cellcolor{green!25}$\mathcal{A}_{\mathrm{dj}}(Vot_{=})$ & 23.33 / 8.13 / 12.06\NEW{$^{+}$} & 45.76 / 19.43 / 27.28\NEW{$^{+}$} \\
& & \cellcolor{green!25}$\mathcal{A}_{\mathrm{dj}}(Vot_{\mathrm{el}})$ & 12.82 / 12.98 / 12.90\NEW{$^{+}$} & 28.01 / 26.64 / 27.31\NEW{$^{+}$} \\
& & $\mathcal{A}_{\mathrm{dj}}(\mathcal{A}_1)$ & 18.31 / 19.72 / \textit{18.99}\NEW{$^{\ast}$} & 41.12 / 39.95 / \textit{40.53}\NEW{$^{\ast}$} \\
\bottomrule
\end{tabular}
}

\end{table}
\textbf{Overview} For the ACOS quadruple extraction task, the results in Table \ref{tab:llm_acos} demonstrate a performance gap that highlights the distinction between learnt pattern recognition and inference-time methods. \CHG{The SFT baselines, particularly \textbf{UnifiedABSA} ($42.58\%$ on Laptop and $60.60\%$ on Restaurant) and \textbf{BART-CRN} ($38.32\%$ and $48.90\%$), maintain a noticeable lead.} Unlike SFT baselines, which are directly optimised on task-specific training data, inference-time adaptation methods (LLM-based Annotators) face an inherent disadvantage: they cannot rely on learned statistical likelihoods to navigate these primary challenges. This gap is further exacerbated by two primary challenges:

\begin{itemize}[noitemsep, topsep=2pt, leftmargin=1.75em]

\item \textbf{Presence of Implicit Opinions and Aspects}: The transition from triplets to quadruples introduces a decision-making layer that goes beyond span extraction; the model must determine whether an aspect or opinion is explicit or implicit and then map it to a specific category.

\item \textbf{Search Space Complexity}: In the Laptop dataset, the taxonomy consists of 112 unique categories. For LLM-based annotation methods, this high-entropy choice space yields the best F1 score of only $18.99\%$ while the Restaurant dataset, featuring only 12 categories, presents a much simpler classification task. Here, proposed LLM-based annotation methods are more effective, with the Med tier achieving a competitive $F1$ score of $40.53\%$.
\end{itemize}

\textbf{LLM-As-ACOS Annotators} The highest performance is achieved in the \textbf{Med} tier. However, performance scaling between the \textbf{Mini} and \textbf{Small} tiers is non-monotonic, as base annotators show a temporary decline in the Small tier. Adjudication consistently outperforms the best individual model ($\mathcal{A}_1$) across all model sizes. Employing $\mathcal{A}_{\mathrm{dj}}(\mathcal{A}_1)$ results in absolute $F_1$ improvements over $\mathcal{A}_1$ on the Laptop and Restaurant datasets: $+3.36 / 1.32\%$ for Mini models, $+1.32 / 8.62\%$ for Small models, and $+3.24 / 12.12\%$ for Medium models.

\NEW{\textbf{Voting vs LLM Adjudication.} Voting and LLM adjudication demonstrate precision-recall trade-offs as with the ASTE task. Exact matching ($Vot_{=}$) serves as a strict precision filter (e.g., $59.36\%$ precision on Restaurant and $32.57\%$ on Laptop at Mini) by excluding non-verbatim quadruples. In contrast, flexible ($Vot_{\approx}$) and element-wise ($Vot_{\mathrm{el}}$) variants improve recall (e.g., $27.18\%$ versus $16.27\%$ on Mini Restaurant). Across all model tiers, LLM-based adjudication consistently outperforms rule-based voting, which is attributed to the increased structural complexity of quadruples:
\begin{itemize}[noitemsep, topsep=2pt, leftmargin=1.5em]
    \item \textbf{Mini Tier:} $\mathcal{A}_{\mathrm{dj}}(\mathcal{A}_1)$ achieves the highest scores on both Laptop ($16.43\%$) and Restaurant ($31.63\%$), while the strongest voting variant ($Vot_{\mathrm{el}}$) attains lower peaks of $13.18\%$ and $28.79\%$, respectively.
    \item \textbf{Small Tier:} In contrast to ASTE, adjudication continues to outperform voting in this tier. $\mathcal{A}_{\mathrm{dj}}(\mathcal{A}_1)$ achieves the highest $F_1$ scores on Laptop ($11.01\%$) and Restaurant ($29.98\%$), surpassing exact voting ($8.15\%$ and $25.12\%$).
    \item \textbf{Medium Tier:} $\mathcal{A}_{\mathrm{dj}}(\mathcal{A}_1)$ achieves the highest overall performance ($18.99\%$ on Laptop and $40.53\%$ on Restaurant), outperforming $Vot_{=}$ ($12.06\%$ and $27.28\%$). The reduced recall of $Vot_{=}$ ($8.13\%$ and $19.43\%$) substantially limits $F_1$ despite its high precision.
\end{itemize}}

\section{Analysis}\label{sec:analysis}

The element-wise and pairwise decomposition of triplet and quadruple evaluations, detailed in Tables \ref{tab:aste_element_assess} and \ref{tab:acos_element_assess}, reveals a consistent pattern across both tasks. Alignment with human annotations is strong at the individual span level but degrades systematically as relational complexity increases. Decomposing the joint evaluation of triplets (ASTE) and quadruples (ACOS) into their constituent labels allows us to identify where this degradation occurs and quantify the relative difficulty of interdependent subtasks.

\begin{table}[ht!]
\setlength{\tabcolsep}{2.25pt}
\centering
\caption{Element-wise assessment of ASTE's alignment with human annotations (F1 score).}
\label{tab:aste_element_assess}
\resizebox{0.85\linewidth}{!}{%
\begin{tabular}{cc cccccc cccccc cccccc}
\toprule
\multirow{2}{*}{\textbf{Data}} & \multirow{2}{*}{$\mathcal{A}$}
  & \multicolumn{6}{c}{Mini}
  & \multicolumn{6}{c}{Small}
  & \multicolumn{6}{c}{Med} \\
\cmidrule(lr){3-8}\cmidrule(lr){9-14}\cmidrule(lr){15-20}
& & $s$ & $at$ & $op$ & $s\&at$ & $s\&op$ & $at\&op$
  & $s$ & $at$ & $op$ & $s\&at$ & $s\&op$ & $at\&op$
  & $s$ & $at$ & $op$ & $s\&at$ & $s\&op$ & $at\&op$ \\
\midrule
\multirow{4}{*}{lap14}
  & $\mathcal{A}_1$ & \heatmap{83.55} & \heatmap{66.74} & \heatmap{64.02} & \heatmap{58.22} & \heatmap{56.53} & \heatmap{48.67} & \heatmap{85.80} & \heatmap{66.18} & \heatmap{65.93} & \heatmap{59.56} & \heatmap{60.36} & \heatmap{48.91} & \heatmap{84.81} & \heatmap{66.12} & \heatmap{72.73} & \heatmap{58.02} & \heatmap{64.64} & \heatmap{51.33} \\
  & $\mathcal{A}_2$ & \heatmap{80.81} & \heatmap{63.21} & \heatmap{50.98} & \heatmap{53.82} & \heatmap{45.61} & \heatmap{39.82} & \heatmap{83.29} & \heatmap{64.30} & \heatmap{61.88} & \heatmap{55.96} & \heatmap{53.97} & \heatmap{47.03} & \heatmap{85.80} & \heatmap{64.46} & \heatmap{68.75} & \heatmap{56.49} & \heatmap{60.52} & \heatmap{49.96} \\
  & $\mathcal{A}_3$ & \heatmap{79.77} & \heatmap{51.34} & \heatmap{37.39} & \heatmap{43.30} & \heatmap{32.01} & \heatmap{26.99} & \heatmap{86.25} & \heatmap{58.32} & \heatmap{49.31} & \heatmap{52.30} & \heatmap{44.12} & \heatmap{36.87} & \heatmap{83.29} & \heatmap{64.30} & \heatmap{61.88} & \heatmap{55.96} & \heatmap{53.97} & \heatmap{47.03} \\

  & $\mathcal{A}dj (Vot_{el})$ &  \heatmap{83.50} & \heatmap{66.46} & \heatmap{61.18} & \heatmap{58.25} & \heatmap{53.88} & \heatmap{46.03} & \heatmap{85.59} & \heatmap{66.18} & \heatmap{61.68} & \heatmap{59.00} & \heatmap{55.36} & \heatmap{47.47} & \heatmap{84.94} & \heatmap{67.28} & \heatmap{72.46} & \heatmap{58.22} & \heatmap{63.98} & \heatmap{52.93} \\
  
  & $\mathcal{A}dj (Vot_{\approx})$ & \heatmap{79.94} & \heatmap{64.55} & \heatmap{51.61} & \heatmap{58.35} & \heatmap{46.68} & \heatmap{43.64} & \heatmap{83.86} & \heatmap{64.73} & \heatmap{57.23} & \heatmap{59.18} & \heatmap{51.97} & \heatmap{47.09} & \heatmap{84.21} & \heatmap{67.45} & \heatmap{67.63} & \heatmap{58.39} & \heatmap{58.76} & \heatmap{51.72} \\

  & $\mathcal{A}dj (Vot_{=})$ & \heatmap{72.22} & \heatmap{60.57} & \heatmap{55.37} & \heatmap{54.57} & \heatmap{50.58} & \heatmap{47.47} & \heatmap{79.94} & \heatmap{64.47} & \heatmap{59.59} & \heatmap{59.15} & \heatmap{54.42} & \heatmap{49.33} & \heatmap{79.88} & \heatmap{65.31} & \heatmap{70.37} & \heatmap{56.63} & \heatmap{61.96} & \heatmap{54.51} \\
  
  & $\mathcal{A}dj (\mathcal{A}_1)$& \heatmap{83.46} & \heatmap{67.38} & \heatmap{69.24} & \heatmap{57.62} & \heatmap{61.55} & \heatmap{52.62} & \heatmap{84.21} & \heatmap{63.81} & \heatmap{60.43} & \heatmap{56.07} & \heatmap{53.82} & \heatmap{45.63} & \heatmap{84.44} & \heatmap{67.36} & \heatmap{72.91} & \heatmap{58.03} & \heatmap{64.37} & \heatmap{53.37} \\
\cmidrule{1-20}
\multirow{4}{*}{res14}
  & $\mathcal{A}_1$ & \heatmap{90.86} & \heatmap{79.93} & \heatmap{75.50} & \heatmap{75.64} & \heatmap{71.77} & \heatmap{67.13} & \heatmap{89.94} & \heatmap{78.35} & \heatmap{77.00} & \heatmap{73.95} & \heatmap{73.06} & \heatmap{66.43} & \heatmap{90.44} & \heatmap{81.35} & \heatmap{79.38} & \heatmap{77.53} & \heatmap{75.73} & \heatmap{71.10} \\
  & $\mathcal{A}_2$ & \heatmap{85.61} & \heatmap{77.74} & \heatmap{68.55} & \heatmap{70.72} & \heatmap{62.91} & \heatmap{61.69} & \heatmap{88.21} & \heatmap{77.08} & \heatmap{72.02} & \heatmap{72.48} & \heatmap{67.76} & \heatmap{62.42} & \heatmap{89.53} & \heatmap{78.65} & \heatmap{75.79} & \heatmap{73.73} & \heatmap{71.70} & \heatmap{65.86} \\
  & $\mathcal{A}_3$ & \heatmap{86.49} & \heatmap{63.91} & \heatmap{53.05} & \heatmap{59.00} & \heatmap{49.63} & \heatmap{40.38} & \heatmap{90.50} & \heatmap{77.81} & \heatmap{68.51} & \heatmap{73.69} & \heatmap{65.24} & \heatmap{59.86} & \heatmap{88.19} & \heatmap{77.35} & \heatmap{72.29} & \heatmap{72.75} & \heatmap{67.97} & \heatmap{62.83} \\

  & $\mathcal{A}dj (Vot_{el})$ &  \heatmap{90.91} & \heatmap{80.50} & \heatmap{74.82} & \heatmap{76.63} & \heatmap{71.10} & \heatmap{66.70} & \heatmap{90.46} & \heatmap{79.09} & \heatmap{75.56} & \heatmap{75.56} & \heatmap{71.78} & \heatmap{66.17} & \heatmap{89.80} & \heatmap{80.74} & \heatmap{77.76} & \heatmap{76.44} & \heatmap{73.92} & \heatmap{69.47} \\
  & $\mathcal{A}dj (Vot_{\approx})$ & \heatmap{89.86} & \heatmap{79.32} & \heatmap{70.16} & \heatmap{76.54} & \heatmap{67.64} & \heatmap{63.01} & \heatmap{90.46} & \heatmap{79.51} & \heatmap{73.72} & \heatmap{76.00} & \heatmap{70.52} & \heatmap{65.99} & \heatmap{90.14} & \heatmap{80.90} & \heatmap{77.70} & \heatmap{76.66} & \heatmap{73.88} & \heatmap{69.68} \\ 
& $\mathcal{A}dj (Vot_{=})$ &\heatmap{86.84} & \heatmap{77.50} & \heatmap{73.20} & \heatmap{75.13} & \heatmap{70.77} & \heatmap{66.33} & \heatmap{89.61} & \heatmap{79.67} & \heatmap{75.10} & \heatmap{76.87} & \heatmap{72.00} & \heatmap{68.15} & \heatmap{89.06} & \heatmap{82.05} & \heatmap{79.04} & \heatmap{77.80} & \heatmap{75.31} & \heatmap{72.07} \\
  &  $\mathcal{A}dj (\mathcal{A}_1)$ & \heatmap{90.63} & \heatmap{80.47} & \heatmap{79.97} & \heatmap{75.97} & \heatmap{76.05} & \heatmap{70.59} & \heatmap{90.24} & \heatmap{78.56} & \heatmap{73.58} & \heatmap{74.11} & \heatmap{69.96} & \heatmap{64.78} & \heatmap{90.20} & \heatmap{81.64} & \heatmap{79.82} & \heatmap{77.28} & \heatmap{75.98} & \heatmap{72.24} \\
\cmidrule{1-20}
\multirow{4}{*}{res15}
  & $\mathcal{A}_1$ & \heatmap{88.49} & \heatmap{76.80} & \heatmap{72.87} & \heatmap{69.92} & \heatmap{64.91} & \heatmap{61.19} & \heatmap{90.53} & \heatmap{82.39} & \heatmap{73.86} & \heatmap{75.62} & \heatmap{67.52} & \heatmap{66.11} & \heatmap{89.50} & \heatmap{77.44} & \heatmap{78.25} & \heatmap{71.00} & \heatmap{71.30} & \heatmap{67.79} \\
  & $\mathcal{A}_2$ & \heatmap{83.26} & \heatmap{75.46} & \heatmap{63.20} & \heatmap{66.19} & \heatmap{54.44} & \heatmap{54.37} & \heatmap{88.71} & \heatmap{78.43} & \heatmap{68.71} & \heatmap{71.80} & \heatmap{61.82} & \heatmap{59.70} & \heatmap{89.47} & \heatmap{78.92} & \heatmap{76.62} & \heatmap{72.42} & \heatmap{69.38} & \heatmap{66.04} \\
  & $\mathcal{A}_3$ & \heatmap{85.96} & \heatmap{60.81} & \heatmap{51.07} & \heatmap{53.82} & \heatmap{46.01} & \heatmap{38.76} & \heatmap{91.01} & \heatmap{79.43} & \heatmap{65.75} & \heatmap{73.85} & \heatmap{59.06} & \heatmap{55.76} & \heatmap{88.74} & \heatmap{78.52} & \heatmap{68.83} & \heatmap{71.92} & \heatmap{61.96} & \heatmap{59.84} \\

  & $\mathcal{A}dj (Vot_{el})$ &  \heatmap{88.52} & \heatmap{76.97} & \heatmap{71.20} & \heatmap{69.11} & \heatmap{63.14} & \heatmap{59.68} & \heatmap{90.15} & \heatmap{82.66} & \heatmap{71.95} & \heatmap{76.15} & \heatmap{64.91} & \heatmap{63.72} & \heatmap{89.10} & \heatmap{78.89} & \heatmap{77.93} & \heatmap{72.71} & \heatmap{70.31} & \heatmap{67.46} \\
  & $\mathcal{A}dj (Vot_{\approx})$ & \heatmap{86.35} & \heatmap{77.27} & \heatmap{67.03} & \heatmap{70.88} & \heatmap{61.14} & \heatmap{57.70} & \heatmap{89.92} & \heatmap{82.63} & \heatmap{70.68} & \heatmap{76.15} & \heatmap{63.86} & \heatmap{63.31} & \heatmap{89.54} & \heatmap{79.38} & \heatmap{77.26} & \heatmap{73.83} & \heatmap{70.22} & \heatmap{66.98} \\
& $\mathcal{A}dj (Vot_{=})$ &\heatmap{82.17} & \heatmap{74.52} & \heatmap{70.48} & \heatmap{69.30} & \heatmap{64.52} & \heatmap{61.63} & \heatmap{85.63} & \heatmap{81.23} & \heatmap{72.51} & \heatmap{74.36} & \heatmap{65.88} & \heatmap{65.79} & \heatmap{87.32} & \heatmap{79.41} & \heatmap{78.70} & \heatmap{73.71} & \heatmap{71.98} & \heatmap{69.55} \\
  & $\mathcal{A}dj (\mathcal{A}_1)$      & \heatmap{88.48} & \heatmap{79.02} & \heatmap{76.14} & \heatmap{71.10} & \heatmap{68.37} & \heatmap{65.13} & \heatmap{89.85} & \heatmap{80.56} & \heatmap{70.76} & \heatmap{73.62} & \heatmap{63.94} & \heatmap{62.63} & \heatmap{89.69} & \heatmap{79.82} & \heatmap{78.27} & \heatmap{73.73} & \heatmap{71.63} & \heatmap{67.99} \\
\cmidrule{1-20}
\multirow{4}{*}{res16}
  & $\mathcal{A}_1$ & \heatmap{88.75} & \heatmap{81.83} & \heatmap{78.93} & \heatmap{76.14} & \heatmap{73.16} & \heatmap{71.23} & \heatmap{89.79} & \heatmap{79.35} & \heatmap{79.61} & \heatmap{73.74} & \heatmap{74.29} & \heatmap{69.48} & \heatmap{88.98} & \heatmap{78.83} & \heatmap{80.55} & \heatmap{73.16} & \heatmap{74.68} & \heatmap{70.14} \\
  & $\mathcal{A}_2$ & \heatmap{88.83} & \heatmap{75.00} & \heatmap{72.00} & \heatmap{69.78} & \heatmap{67.05} & \heatmap{62.69} & \heatmap{87.14} & \heatmap{76.36} & \heatmap{75.34} & \heatmap{70.04} & \heatmap{69.56} & \heatmap{63.49} & \heatmap{90.91} & \heatmap{76.98} & \heatmap{78.16} & \heatmap{72.18} & \heatmap{73.35} & \heatmap{66.79} \\
  & $\mathcal{A}_3$ & \heatmap{87.91} & \heatmap{65.34} & \heatmap{57.80} & \heatmap{61.07} & \heatmap{53.83} & \heatmap{47.58} & \heatmap{89.97} & \heatmap{79.01} & \heatmap{74.42} & \heatmap{74.11} & \heatmap{70.54} & \heatmap{65.83} & \heatmap{87.14} & \heatmap{76.36} & \heatmap{75.34} & \heatmap{70.04} & \heatmap{69.56} & \heatmap{63.49} \\

  & $\mathcal{A}dj (Vot_{el})$ &  \heatmap{90.60} & \heatmap{83.09} & \heatmap{77.79} & \heatmap{78.08} & \heatmap{72.85} & \heatmap{71.31} & \heatmap{89.26} & \heatmap{79.88} & \heatmap{78.40} & \heatmap{74.28} & \heatmap{73.31} & \heatmap{68.67} & \heatmap{88.95} & \heatmap{77.56} & \heatmap{79.33} & \heatmap{71.54} & \heatmap{73.46} & \heatmap{67.20} \\
  & $\mathcal{A}dj (Vot_{\approx})$ & \heatmap{90.50} & \heatmap{81.51} & \heatmap{75.88} & \heatmap{78.28} & \heatmap{72.97} & \heatmap{69.35} & \heatmap{89.39} & \heatmap{81.40} & \heatmap{78.39} & \heatmap{76.00} & \heatmap{73.17} & \heatmap{69.71} & \heatmap{89.55} & \heatmap{78.65} & \heatmap{79.60} & \heatmap{73.13} & \heatmap{74.40} & \heatmap{68.17} \\
& $\mathcal{A}dj (Vot_{=})$ & \heatmap{88.30} & \heatmap{79.81} & \heatmap{78.92} & \heatmap{77.46} & \heatmap{76.46} & \heatmap{72.97} & \heatmap{88.63} & \heatmap{80.77} & \heatmap{79.96} & \heatmap{75.77} & \heatmap{74.92} & \heatmap{71.64} & \heatmap{88.67} & \heatmap{78.46} & \heatmap{79.47} & \heatmap{73.12} & \heatmap{74.13} & \heatmap{69.07} \\
  & $\mathcal{A}dj (\mathcal{A}_1)$ & \heatmap{89.55} & \heatmap{81.96} & \heatmap{81.33} & \heatmap{76.29} & \heatmap{75.70} & \heatmap{72.49} & \heatmap{89.04} & \heatmap{78.09} & \heatmap{77.21} & \heatmap{72.44} & \heatmap{71.77} & \heatmap{67.08} & \heatmap{89.73} & \heatmap{80.04} & \heatmap{82.44} & \heatmap{74.62} & \heatmap{76.45} & \heatmap{72.17} \\
\bottomrule
\end{tabular}}

\end{table}

\begin{table}[htbp]
\setlength{\tabcolsep}{3pt}
\centering
\caption{Element-wise assessment of ACOS alignment with human annotations (F1 score). Complete table can be found in Appendix \ref{sec:complete_acos}}
\label{tab:acos_element_assess}
\resizebox{0.75\linewidth}{!}{%
\begin{tabular}{c | c c | c c c c c c c c c c c c c}
\toprule
Dataset
  & \multicolumn{2}{c|}{Scale \& Annot}
  & $s$ & $at$ & $op$ & $E$ & $A$ & $ac$
  & \shortstack{$s$\\+\\$at$}
  & \shortstack{$s$\\+\\$op$}
  & \shortstack{$at$\\+\\$op$}
  & \shortstack{$s$\\+\\$ac$}
  & \shortstack{$at$\\+\\$ac$}
  & \shortstack{$op$\\+\\$ac$}
  & \shortstack{$at$+\\$s$+\\$op$} \\
\midrule
\multirow{4}{*}{laptop}
  & \multirow{13}{*}{\rotatebox{90}{med}} & $\mathcal{A}_1$ & \heatmap{89.86} & \heatmap{68.20} & \heatmap{55.29} & \heatmap{62.96} & \heatmap{53.05} & \heatmap{36.30} & \heatmap{62.67} & \heatmap{52.17} & \heatmap{39.57} & \heatmap{33.35} & \heatmap{25.95} & \heatmap{22.61} & \heatmap{37.46} \\
  & & $\mathcal{A}_2$ & \heatmap{90.37} & \heatmap{61.84} & \heatmap{43.96} & \heatmap{56.42} & \heatmap{56.75} & \heatmap{33.15} & \heatmap{57.22} & \heatmap{42.07} & \heatmap{29.04} & \heatmap{30.59} & \heatmap{20.72} & \heatmap{15.40} & \heatmap{27.66} \\
  & & $\mathcal{A}_3$ & \heatmap{89.30} & \heatmap{57.66} & \heatmap{44.10} & \heatmap{49.61} & \heatmap{49.31} & \heatmap{21.97} & \heatmap{53.06} & \heatmap{41.23} & \heatmap{27.21} & \heatmap{19.92} & \heatmap{15.54} & \heatmap{10.62} & \heatmap{25.65} \\

  & &  $\mathcal{A}dj (Vot_{el})$ & \heatmap{90.17} & \heatmap{66.60} & \heatmap{50.05} & \heatmap{61.46} & \heatmap{53.97} & \heatmap{34.87} & \heatmap{61.89} & \heatmap{47.46} & \heatmap{35.71} & \heatmap{32.28} & \heatmap{24.09} & \heatmap{18.71} & \heatmap{33.79} \\
  & & $\mathcal{A}dj(V_{\approx})$& \heatmap{86.19} & \heatmap{65.47} & \heatmap{39.38} & \heatmap{57.09} & \heatmap{50.11} & \heatmap{31.39} & \heatmap{61.07} & \heatmap{37.62} & \heatmap{29.55} & \heatmap{29.57} & \heatmap{23.28} & \heatmap{15.89} & \heatmap{28.04} \\

   & & $\mathcal{A}dj(V_{=})$& \heatmap{54.41} & \heatmap{44.31} & \heatmap{30.08} & \heatmap{38.64} & \heatmap{35.85} & \heatmap{24.22} & \heatmap{41.90} & \heatmap{28.90} & \heatmap{24.34} & \heatmap{22.88} & \heatmap{20.60} & \heatmap{14.42} & \heatmap{23.42} \\
  
  & & $\mathcal{A}dj(\mathcal{A}_1)$ & \heatmap{90.40} & \heatmap{73.53} & \heatmap{59.59} & \heatmap{70.51} & \heatmap{54.27} & \heatmap{41.23} & \heatmap{67.71} & \heatmap{56.60} & \heatmap{45.80} & \heatmap{37.90} & \heatmap{30.68} & \heatmap{24.84} & \heatmap{43.53} \\
\cmidrule{1-1}\cmidrule{3-16}
\multirow{4}{*}{restaurant}
  & & $\mathcal{A}_1$ & \heatmap{89.94} & \heatmap{69.39} & \heatmap{64.13} & \heatmap{65.41} & \heatmap{72.28} & \heatmap{54.91} & \heatmap{63.90} & \heatmap{60.16} & \heatmap{50.09} & \heatmap{49.01} & \heatmap{39.26} & \heatmap{37.66} & \heatmap{47.46} \\
  & & $\mathcal{A}_2$ & \heatmap{91.50} & \heatmap{63.30} & \heatmap{54.71} & \heatmap{52.97} & \heatmap{63.72} & \heatmap{46.25} & \heatmap{58.98} & \heatmap{52.52} & \heatmap{39.54} & \heatmap{42.48} & \heatmap{28.08} & \heatmap{25.09} & \heatmap{38.32} \\
  & & $\mathcal{A}_3$ & \heatmap{87.93} & \heatmap{60.00} & \heatmap{49.27} & \heatmap{54.07} & \heatmap{68.22} & \heatmap{49.11} & \heatmap{56.03} & \heatmap{46.46} & \heatmap{37.32} & \heatmap{44.55} & \heatmap{36.23} & \heatmap{27.91} & \heatmap{35.69} \\

   & &  $\mathcal{A}dj (Vot_{el})$ & \heatmap{91.08} & \heatmap{69.29} & \heatmap{60.44} & \heatmap{65.23} & \heatmap{73.31} & \heatmap{56.22} & \heatmap{64.37} & \heatmap{57.55} & \heatmap{47.90} & \heatmap{51.58} & \heatmap{39.25} & \heatmap{36.29} & \heatmap{45.87} \\

  & & $\mathcal{A}dj(V_{\approx})$ & \heatmap{87.88} & \heatmap{67.69} & \heatmap{51.52} & \heatmap{59.43} & \heatmap{66.00} & \heatmap{49.67} & \heatmap{63.96} & \heatmap{49.90} & \heatmap{42.45} & \heatmap{46.14} & \heatmap{37.44} & \heatmap{29.14} & \heatmap{41.10} \\

  & & $\mathcal{A}dj(V_{=})$& \heatmap{63.03} & \heatmap{52.60} & \heatmap{46.53} & \heatmap{47.01} & \heatmap{50.73} & \heatmap{42.66} & \heatmap{49.71} & \heatmap{45.48} & \heatmap{39.41} & \heatmap{39.74} & \heatmap{36.29} & \heatmap{31.48} & \heatmap{38.47} \\
  
  & &  $\mathcal{A}dj(\mathcal{A}_1)$ & \heatmap{91.01} & \heatmap{72.74} & \heatmap{64.60} & \heatmap{82.96} & \heatmap{82.43} & \heatmap{71.75} & \heatmap{68.13} & \heatmap{61.17} & \heatmap{52.86} & \heatmap{66.40} & \heatmap{55.78} & \heatmap{48.85} & \heatmap{50.06} \\
\bottomrule
\end{tabular}}

\end{table}

\textbf{Performance Hierarchy and Label Alignment}:  Across both tasks, sentiment polarity ($s$) consistently \CHG{exhibits} the highest alignment with human annotations, \CHG{approaching, and in some settings exceeding,} 90\% F1. Aspect term ($at$) and opinion term ($op$) spans show lower, yet promising, alignment scores. Performance declines predictably as models are required to identify interdependent structures. In the ASTE task (Table \ref{tab:aste_element_assess}), while assigning the correct polarity to an aspect ($s\&at$) is more challenging than individual labelling, the most degrading bottleneck appears to be $at\&op$ identification, i.e. the process of correctly linking an opinion span to its specific target.

\textbf{Domain Disparity and Complexity in ACOS} Similar trends persist in the ACOS task (Table \ref{tab:acos_element_assess}), though the increased dimensionality introduces further complexity. Models exhibit a pronounced domain disparity, particularly in the laptop domain, where the lack of dataset-specific tuning appears to hinder performance.  This difficulty is largely due to the prediction of aspect categories (ac), which reinforces the previously discussed challenge of increased search-space complexity for $ac$, with 112 categories in the Laptop domain and 12 in the restaurant domain. The lower scores for the aspect term and opinion span than for the ASTE counterparts can be explained by the presence of implicit sentiments and aspects, which must be left unlabelled (or set to None) when labelling quadruples.

\textbf{Adjudication Effects} \CHG{The LLM adjudicator ($\mathcal{A}_{\mathrm{dj}}(\mathcal{A}_1)$) reduces single-annotator noise and improves both paired and combined evaluation scores in most scenarios. For example, $at \& op$ alignment on lap14 Mini increases from 48.67\% ($\mathcal{A}_1$) to 52.62\%, and $op + ac$ on Restaurant Med rises from 37.66\% to 48.85\%. The only exception is observed in the small-scale tier (discussed previously in Table \ref{tab:small_adj_ablation}). A comparison of consensus methods reveals distinct trade-offs. Element-wise ($Vot_{\mathrm{el}}$) and flexible ($Vot_{\approx}$) voting approaches maintain baseline performance on individual spans but provide limited improvement for interdependent combinations. In contrast, exact voting ($Vot_{=}$) performs poorly on high-dimensional tasks, with its strict agreement threshold reducing Laptop Med sentiment alignment from 89.86\% under $\mathcal{A}_1$ to 54.41\%. Overall, LLM-based adjudication is particularly effective at resolving fine-grained relational dependencies, an area where rule-based consensus methods are less successful.
} 

\begin{figure}
    \centering
    \includegraphics[width=0.95\linewidth]{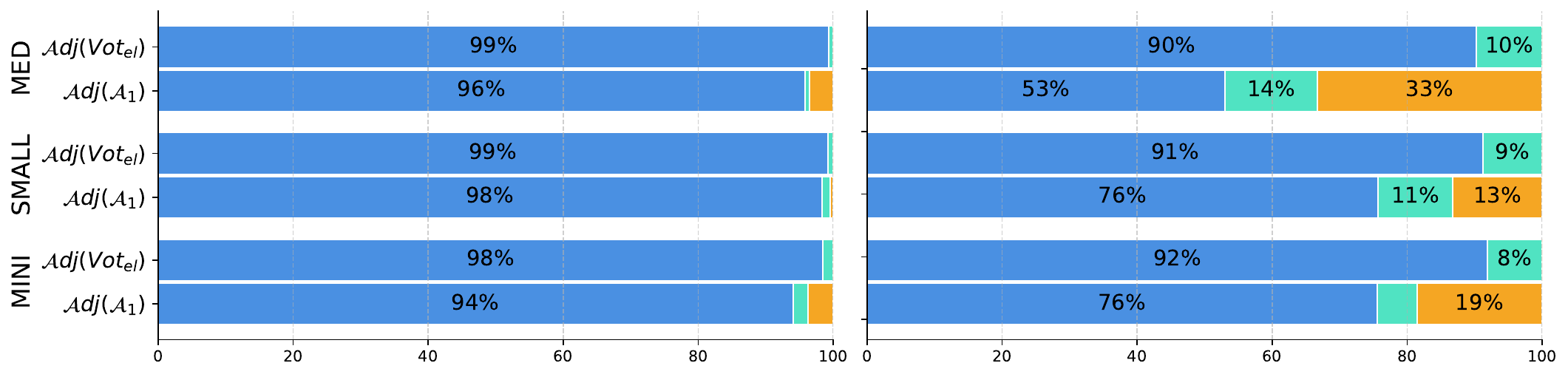}
    \caption{\textcolor{addgreen}{Output composition for LLM-Adjudication ($\mathcal{A}dj(\mathcal{A}_1)$) and Element-Wise adjudicators ($\mathcal{A}dj(Vot_{el})$). Highlighting:} \colorbox{colcopy}{\textit{copy}}, \colorbox{colrecomb}{\textit{recombination}} \NEW{and} \colorbox{colgen}{\textit{generation}} \NEW{observed.}} 
\label{fig:mechanism}
\end{figure}

\NEW{\subsection{Adjudication Mechanism}\label{sec:mechanism}
To explain the outcome above, we decompose how each aggregation method produces its output. Each emitted tuple is classified as: (i) \emph{copy}: identical to an annotator tuple, (ii) \emph{recombination}: not a copy, but every element appears among the candidates, or (iii) \emph{generation}: contains an element not proposed by any annotator. Based on the underlying mechanism, there are idiosyncrasies: (a) exact and flexible voting are limited to copying, (b) element-wise voting enables recombination, and (c) the LLM adjudicator can generate, as it is the sole method that re-examines the source text. Figure \ref{fig:mechanism} demonstrates that ASTE outputs consist of $94\%–98\%$ copies. In comparison, the higher-dimensional ACOS task reduces simple copying, resulting in $14\%$ recombination and $33\%$ generation, with $53\%$ copying at the Med scale. Conversely, Element-Wise voting remains predominantly limited to copies, achieving a maximum of $10\%$ recombination (refer to Appendix \ref{sec:mechfull} for more details).}

\subsection{Inter-Annotator Agreement of LLM-Based Annotators}
\begin{table}[thbp]
\centering
\caption{Top: Krippendorff-$\alpha$ for IAA for ASTE span annotations ($at$: Aspect, $op$: Opinion, $at \& s$: Aspect Term with sentiment, $op \& s$: Opinion span with sentiment). Bottom: Performance scores for ACOS tasks ($ac$: Aspect Category).}
\label{tab:combined_tables}
\setlength{\tabcolsep}{3pt}
\resizebox{0.815\linewidth}{!}{
\begin{tabular}{l c c c c c c c c c c c c c c c c}
\toprule
\multirow{2}{*}{Model} & \multicolumn{4}{c}{lap14} & \multicolumn{4}{c}{res14} & \multicolumn{4}{c}{res15} & \multicolumn{4}{c}{res16} \\
\cmidrule(lr){2-5} \cmidrule(lr){6-9} \cmidrule(lr){10-13} \cmidrule(lr){14-17}
 & $at$ & $op$ & $at \& s$ & $op \& s$ & $at$ & $op$ & $at \& s$ & $op \& s$ & $at$ & $op$ & $at \& s$ & $op \& s$ & $at$ & $op$ & $at \& s$ & $op \& s$ \\
\midrule
Mini  & 0.6125 & \cellcolor{lowRelRed} 0.49 & \cellcolor{lowRelRed} 0.59 & \cellcolor{lowRelRed} 0.46 & 0.70 & \cellcolor{lowRelRed} 0.61 & 0.68 & \cellcolor{lowRelRed} 0.59 & 0.70 &  \cellcolor{lowRelRed} 0.57 & 0.67 & \cellcolor{lowRelRed} 0.54 & 0.72 & \cellcolor{lowRelRed} 0.63 & 0.69 & \cellcolor{lowRelRed} 0.60 \\
Small & 0.73 &  0.66 & 0.72 &  0.66 & 0.79 & 0.74 & 0.78 & 0.74 & \cellcolor{highRelBlue}0.81 & 0.72 & \cellcolor{highRelBlue} 0.80 & 0.71 & 0.79 & 0.77 & 0.78 & 0.75 \\
Med   & 0.75 & 0.69 & 0.75 & 0.69 & \cellcolor{highRelBlue} 0.82 & 0.77 & \cellcolor{highRelBlue}0.80 & 0.77 & \cellcolor{highRelBlue} 0.84 & 0.75 & \cellcolor{highRelBlue} 0.82 & 0.74 & \cellcolor{highRelBlue} 0.82 & 0.79 & \cellcolor{highRelBlue} 0.81 & 0.78 \\
\bottomrule
\vspace{3pt}
\end{tabular}
}

\resizebox{0.55\linewidth}{!}{
\begin{tabular}{l c c c c c c c c c c}
\toprule
\multirow{2}{*}{Model} & \multicolumn{5}{c}{Laptop} & \multicolumn{5}{c}{Restaurant} \\
\cmidrule(lr){2-6} \cmidrule(lr){7-11}
 & $at$ & $op$ & $ac$ & $at \& s$ & $op \& s$ & $at$ & $op$ & $ac$ & $at \& s$ & $op \& s$ \\
\midrule
Mini  & \cellcolor{lowRelRed} 0.60 & \cellcolor{lowRelRed}0.37 & \cellcolor{lowRelRed}0.22 & \cellcolor{lowRelRed} 0.58 & \cellcolor{lowRelRed}0.37 & \cellcolor{lowRelRed}0.61 & \cellcolor{lowRelRed}0.41 & \cellcolor{lowRelRed}0.41 & \cellcolor{lowRelRed}0.59 & \cellcolor{lowRelRed}0.40 \\
Small & 0.70 & \cellcolor{lowRelRed}0.65 & \cellcolor{lowRelRed}0.31 & 0.69 & \cellcolor{lowRelRed}0.65 & 0.73 & \cellcolor{lowRelRed}0.64 & \cellcolor{lowRelRed}0.49 & 0.71 & \cellcolor{lowRelRed}0.64 \\
Med   & 0.70 & \cellcolor{lowRelRed}0.64 & \cellcolor{lowRelRed}0.33 & 0.69 & \cellcolor{lowRelRed}0.64 & 0.71 & 0.65 & \cellcolor{lowRelRed}0.50 & 0.69 & \cellcolor{lowRelRed}0.64 \\
\bottomrule
\end{tabular}
}

\end{table}
While joint evaluation shows promising ASTE results (Table~\ref{tab:llm_aste_trip}), ACOS performance (Table~\ref{tab:llm_acos}) confirms that LLMs cannot yet perform complex annotation independently. However, element-wise analysis (Tables~\ref{tab:aste_element_assess} and~\ref{tab:acos_element_assess}) reveals strong span-level potential. To rigorously evaluate this reliability beyond standard F1-scores \citep{datainwild_2024}, we report pre-adjudication Inter-Annotator Agreement (IAA) using Krippendorff's $\alpha$. These agreement metrics reveal that reliability for ASTE increases with model parameter size, with triplet span elements approaching \textbf{highly reliable} levels. Conversely, this trend is notably weaker for ACOS, where quadruple elements plateau at \textbf{moderately reliable} levels at best.

\section{Error Analysis}
\begin{table}[!ht]
\centering
\scriptsize
\caption{Qualitative error analysis of ACOS cells display the tuple ($s$: sentiment, $at$: aspect term, $ac$: aspect category, $op$: opinion span). ASTE rows omit the category field. NULL indicates an implicit element. \err{Incorrect} and \err{\underline{hallucinated}} entries are highlighted relative to the Gold annotation.}
\label{tab:acos_qual}
\setlength{\tabcolsep}{3.5pt}
\resizebox{0.85\linewidth}{!}{
\begin{tabular}{p{3cm}|p{3cm}|p{3cm}|p{3cm}}
\toprule
\multicolumn{4}{p{12cm}}{\textbf{[ACOS] Text 1:} \textit{this computer is great for playing videos in a classroom setting of 10 people , connecting to an hdmi tv to show videos and powerpoints , and has a fantastic battery life .}} \\
\midrule

\textbf{Gold:} & \textbf{$\mathcal{A}_1$:} & \textbf{$\mathcal{A}_2$:} & \textbf{$\mathcal{A}_3$:}  \\

\begin{tabular}{@{}l@{}}
$s$: Positive\\$at$: computer\\$ac$: laptop\#op\_perf\\$op$: great \\
\rule{0pt}{3ex}$s$: Positive\\$at$: battery life\\$ac$: battery\#quality\\$op$: fantastic
\end{tabular} &

\begin{tabular}{@{}l@{}}
$s$: Positive\\$at$: computer\\$ac$: \err{lap\#general}\\$op$: great \\
\rule{0pt}{3ex}$s$: Positive\\$at$: battery life\\$ac$: \err{lap\#portability}\\$op$: fantastic
\end{tabular} &

\begin{tabular}{@{}l@{}}
$s$: Positive\\$at$: computer\\$ac$: \err{mul\_device\#op\_perf}\\$op$: great \\
\rule{0pt}{3ex}$s$: Positive\\$at$: battery life\\$ac$: \err{lap\#portability}\\$op$: fantastic\\
\rule{0pt}{3ex}$s$: Positive\\$at$: computer\\$ac$: \err{lap\#connectivity}\\$op$: great
\end{tabular} &

\begin{tabular}{@{}l@{}}
$s$: Positive\\$at$: computer\\$ac$: \err{cpu\#general}\\$op$: great \\
\rule{0pt}{3ex}$s$: Positive\\$at$: \err{hdmi tv}\\$ac$: \err{port\#connectivity}\\$op$: \err{NULL}\\
\rule{0pt}{3ex}$s$: Positive\\$at$: battery life\\$ac$: battery\#quality\\$op$: fantastic
\end{tabular} \\

\midrule

$\mathcal{A}dj (Vot_{el})$: & $\mathcal{A}dj(V_{\approx})$:  & $\mathcal{A}dj(V_{=})$: & $\mathcal{A}dj(\mathcal{A}_1)$: \\

\begin{tabular}{@{}l@{}}
$s$: Positive\\$at$: computer\\$ac$: \err{lap\#general}\\$op$: great \\
\rule{0pt}{3ex}$s$: Positive\\$at$: battery life\\$ac$: \err{lap\#portability}\\$op$: fantastic\\
\end{tabular} &

\begin{tabular}{@{}l@{}}
$s$: Positive\\$at$: computer\\$ac$: \err{lap\#general}\\$op$: great \\
\rule{0pt}{3ex}$s$: Positive\\$at$: battery life\\$ac$: \err{lap\#portability}\\$op$: fantastic\\
\end{tabular} &

\begin{tabular}{@{}l@{}}
\\

\end{tabular} &

\begin{tabular}{@{}l@{}}
$s$: Positive\\$at$: computer\\$ac$: lap\#op\_perf\\$op$: great \\
\rule{0pt}{3ex}$s$: Positive\\$at$: battery life\\$ac$: battery\#quality\\$op$: fantastic
\end{tabular} \\

\toprule
\multicolumn{4}{p{12cm}}{\textbf{[ASTE] Text 1:} \textit{It arrived so fast and customer service was great .}} \\
\midrule
\textbf{Gold:} & \textbf{$\mathcal{A}_1$:} & \textbf{$\mathcal{A}_2$:} & \textbf{$\mathcal{A}_3$:}  \\

\begin{tabular}{@{}l@{}}
$s$: Positive\\$at$: customer service\\$op$: great \\

\end{tabular} &

\begin{tabular}{@{}l@{}}
$s$: Positive\\$at$: \err{\underline{delivery}}\\$op$: \err{fast} \\
\rule{0pt}{3ex}$s$: Positive\\$at$: customer service\\$op$: great \\

\end{tabular} &

\begin{tabular}{@{}l@{}}
$s$: Positive\\$at$: \err{\underline{arrival}}\\$op$: \err{fast} \\
\rule{0pt}{3ex}$s$: Positive\\$at$: customer service\\$op$: great \\

\end{tabular} &

\begin{tabular}{@{}l@{}}
$s$: Positive\\$at$: customer service\\$op$: great \\
\rule{0pt}{3ex}$s$: \err{Negative}\\$at$: \err{\underline{price}}\\$op$: \err{arrived so fast} \\
\end{tabular} \\
\midrule

$\mathcal{A}dj (Vot_{el})$: & $\mathcal{A}dj(V_{\approx})$:  & $\mathcal{A}dj(V_{=})$: & $\mathcal{A}dj(\mathcal{A}_1)$: \\

\begin{tabular}{@{}l@{}}
$s$: Positive\\$at$: \err{\underline{delivery}}\\$op$: fast \\
\rule{0pt}{3ex}$s$: Positive\\$at$: customer service\\$op$: great \\
\end{tabular} &

\begin{tabular}{@{}l@{}}
$s$: Positive\\$at$: customer service\\$op$: great \\
\end{tabular} &

\begin{tabular}{@{}l@{}}
$s$: Positive\\$at$: customer service\\$op$: great \\
\end{tabular} &

\begin{tabular}{@{}l@{}}
$s$: Positive\\$at$: \err{\underline{delivery}}\\$op$: fast \\
\rule{0pt}{3ex}$s$: Positive\\$at$: customer service\\$op$: great \\
\end{tabular}  \\
\bottomrule

\end{tabular}

}

\end{table}

\NEW{Failure modes fall into four categories: (i) opinion--target mislinking ($at\&op$), (ii) category misassignment ($ac$), (iii) implicit-element errors, and (iv) span-boundary divergence. Unlike tuple-level prescribed metrics that obscure these granular errors, element-reorganising methods expose them (Tables~\ref{tab:aste_element_assess}, \ref{tab:acos_element_assess}).}

\CHG{Table 9 highlights how inherent task subjectivity and framing drive these failure modes. In ACOS, plausible category shifts (e.g., Gold's \textit{laptop\#op\_perf} vs $A_1$'s \textit{lap\#general}) are partial misalignments that Exact Match F1 conflates with total failure. Similarly in ASTE (\textit{``It arrived so fast...''}), annotators extract implicit targets like \textit{``delivery''} or \textit{``arrival''} that are absent from the text; while semantically accurate, these violate strict explicit-span conventions and are penalised as hallucinations. This reveals both the promise and the limitations of LLM-based adjudication. On one hand, its ability to recombine and generate elements allows it to capture human labelling dynamics better than rigid voting mechanisms. On the other hand, this flexibility becomes a drawback when task guidelines strictly forbid extracting terms anchored outside the text.}

\section{Conclusion}
This work introduces an LLM-based declarative opinion-annotation pipeline and an opinion-adjudication method to resolve inter-annotator disagreements. The pipelines use ASTE and ACOS schemas to enforce extractive constraints during extraction and adjudication. Evaluation across multiple model scales against human-annotated ground truth reveals a critical performance bifurcation: although LLMs are effective at detecting individual spans, their structural reliability declines markedly as relational complexity increases. \NEW{We benchmark the LLM-based adjudicator against exact, flexible, and element-wise rule-based voting. While exact voting provides a strong high-precision consensus for ASTE triplets, only LLM-based adjudication consistently improves relationally complex ACOS quadruples due to its capability of re-examining the source text to generate missing elements}. The optimal adjudication strategy is therefore a function of relational complexity. Calculating Krippendorff's $\alpha$ among LLM annotators confirms strong span-level utility but weaker reliability in annotating the interdependent structures connecting those spans. At current capability levels, the proposed LLM-based pipeline provides reliable span and span-sentiment annotations, motivating their use as high-fidelity pre-annotation assistants and scalable data augmentation tools rather than stand-alone replacements for human annotators.

\section{Limitation}
Several limitations warrant consideration, starting with our use of model quantisation to deploy large-scale models on consumer-grade hardware, specifically the $32$B variants. While this approach ensures computational efficiency, it may introduce subtle reasoning trade-offs that increase friction in faithfully reproducing relational structures. Secondly, the pipeline's declarative method with inference-time adaptation is theoretically domain-agnostic. However, our current assessment is limited to SemEval-derived review datasets from the laptop and restaurant domains. Consequently, the pipeline’s robustness on more complex or novel unannotated corpora remains a subject for future research. Finally, the process is not strictly annotation-free, as it relies on a small number of in-context learning examples ($k \in \{5, 10, 15\}$). While our method is resource-efficient as it does not require fine-tuning or parametrised optimisation, it still requires a modest foundation of high-quality human labels (at least 5-15 examples) to automatically synthesise prompts.
\bibliography{main}
\bibliographystyle{tmlr}
\newpage
\appendix
\section{Appendix}\label{sec:complete_acos} 
\begin{table}[ht!]
\setlength{\tabcolsep}{3pt}
\centering
\caption{FULL Element-wise assessment of ACOS alignment with human annotations (F1 score).}
\label{tab:acos_element_assess_complete}
\resizebox{0.9\linewidth}{!}{%
\begin{tabular}{c | c c | c c c c c c c c c c c c c}
\toprule
Dataset
  & \multicolumn{2}{c|}{Scale \& Annot}
  & $s$ & $at$ & $op$ & $E$ & $A$ & $ac$
  & \shortstack{$s$\\+\\$at$}
  & \shortstack{$s$\\+\\$op$}
  & \shortstack{$at$\\+\\$op$}
  & \shortstack{$s$\\+\\$ac$}
  & \shortstack{$at$\\+\\$ac$}
  & \shortstack{$op$\\+\\$ac$}
  & \shortstack{$at$+\\$s$+\\$op$} \\
\midrule
\multirow{4}{*}{laptop}
  & \multirow{13}{*}{\rotatebox{90}{mini}} & $\mathcal{A}_1$ & \heatmap{89.48} & \heatmap{68.97} & \heatmap{57.05} & \heatmap{53.05} & \heatmap{54.05} & \heatmap{30.75} & \heatmap{63.47} & \heatmap{54.96} & \heatmap{42.50} & \heatmap{28.26} & \heatmap{21.68} & \heatmap{18.06} & \heatmap{40.65} \\
  & & $\mathcal{A}_2$ & \heatmap{88.52} & \heatmap{62.30} & \heatmap{45.09} & \heatmap{66.95} & \heatmap{54.12} & \heatmap{34.99} & \heatmap{56.61} & \heatmap{42.59} & \heatmap{30.39} & \heatmap{31.43} & \heatmap{23.54} & \heatmap{18.19} & \heatmap{28.55} \\
  & & $\mathcal{A}_3$ & \heatmap{73.39} & \heatmap{37.60} & \heatmap{17.16} & \heatmap{46.63} & \heatmap{39.58} & \heatmap{19.53} & \heatmap{32.25} & \heatmap{15.98} & \heatmap{10.42} & \heatmap{16.43} & \heatmap{7.24}  & \heatmap{3.40}  & \heatmap{9.91}  \\

  &   &  $\mathcal{A}dj (Vot_{el})$  & \heatmap{89.09} & \heatmap{67.76} & \heatmap{52.39} & \heatmap{58.32} & \heatmap{53.41} & \heatmap{32.69} & \heatmap{62.51} & \heatmap{50.23} & \heatmap{38.70} & \heatmap{30.19} & \heatmap{22.98} & \heatmap{18.49} & \heatmap{36.94} \\
  & & $\mathcal{A}dj (V_{\approx})$ & \heatmap{82.05} & \heatmap{61.69} & \heatmap{32.91} & \heatmap{52.45} & \heatmap{44.75} & \heatmap{26.65} & \heatmap{58.15} & \heatmap{32.10} & \heatmap{26.62} & \heatmap{25.04} & \heatmap{19.41} & \heatmap{11.42} & \heatmap{25.90} \\
  & & $\mathcal{A}dj (V_{=})$ & \heatmap{34.62} & \heatmap{27.74} & \heatmap{24.02} & \heatmap{27.66} & \heatmap{26.06} & \heatmap{18.78} & \heatmap{26.40} & \heatmap{23.00} & \heatmap{19.74} & \heatmap{17.83} & \heatmap{15.59} & \heatmap{12.35} & \heatmap{18.85} \\
  & & $\mathcal{A}dj(\mathcal{A}_1)$             & \heatmap{89.70} & \heatmap{68.45} & \heatmap{55.21} & \heatmap{67.81} & \heatmap{54.83} & \heatmap{39.87} & \heatmap{63.12} & \heatmap{52.30} & \heatmap{40.37} & \heatmap{36.29} & \heatmap{27.76} & \heatmap{22.75} & \heatmap{38.50} \\
\cmidrule{1-1}\cmidrule{3-16}
\multirow{4}{*}{Restaurant}
  & & $\mathcal{A}_1$ & \heatmap{90.34} & \heatmap{68.69} & \heatmap{65.58} & \heatmap{62.51} & \heatmap{76.59} & \heatmap{55.87} & \heatmap{63.70} & \heatmap{61.79} & \heatmap{51.20} & \heatmap{51.85} & \heatmap{40.53} & \heatmap{39.00} & \heatmap{48.75} \\
  & & $\mathcal{A}_2$ & \heatmap{87.77} & \heatmap{61.60} & \heatmap{55.47} & \heatmap{57.80} & \heatmap{65.34} & \heatmap{49.39} & \heatmap{58.54} & \heatmap{52.76} & \heatmap{41.05} & \heatmap{45.86} & \heatmap{31.35} & \heatmap{29.70} & \heatmap{39.51} \\
  & & $\mathcal{A}_3$ & \heatmap{76.86} & \heatmap{38.22} & \heatmap{23.84} & \heatmap{34.68} & \heatmap{49.03} & \heatmap{26.51} & \heatmap{34.30} & \heatmap{21.89} & \heatmap{15.26} & \heatmap{23.54} & \heatmap{13.85} & \heatmap{8.46}  & \heatmap{14.27} \\

  &   &  $\mathcal{A}dj (Vot_{el})$ & \heatmap{88.83} & \heatmap{67.82} & \heatmap{62.41} & \heatmap{63.37} & \heatmap{72.93} & \heatmap{55.27} & \heatmap{63.53} & \heatmap{59.85} & \heatmap{49.01} & \heatmap{52.04} & \heatmap{39.68} & \heatmap{37.80} & \heatmap{47.37} \\
  & & $\mathcal{A}dj (V_{\approx})$ & \heatmap{79.67} & \heatmap{60.09} & \heatmap{48.01} & \heatmap{55.12} & \heatmap{63.49} & \heatmap{48.42} & \heatmap{57.66} & \heatmap{46.39} & \heatmap{39.68} & \heatmap{46.57} & \heatmap{36.27} & \heatmap{31.84} & \heatmap{38.61} \\  

    & & $\mathcal{A}dj (V_{=})$ & \heatmap{47.22} & \heatmap{37.26} & \heatmap{38.32} & \heatmap{40.88} & \heatmap{41.27} & \heatmap{36.48} & \heatmap{35.81} & \heatmap{37.33} & \heatmap{31.91} & \heatmap{35.16} & \heatmap{29.26} & \heatmap{30.95} & \heatmap{31.39} \\
  
  & & $\mathcal{A}dj (\mathcal{A}_1)$             & \heatmap{90.48} & \heatmap{69.26} & \heatmap{58.19} & \heatmap{85.88} & \heatmap{75.92} & \heatmap{66.25} & \heatmap{64.77} & \heatmap{54.08} & \heatmap{45.64} & \heatmap{61.83} & \heatmap{50.14} & \heatmap{40.86} & \heatmap{43.00} \\
\midrule
\multirow{4}{*}{laptop}
  & \multirow{13}{*}{\rotatebox{90}{small}} & $\mathcal{A}_1$ & \heatmap{90.88} & \heatmap{59.26} & \heatmap{36.86} & \heatmap{65.58} & \heatmap{59.28} & \heatmap{40.56} & \heatmap{55.24} & \heatmap{35.22} & \heatmap{22.85} & \heatmap{37.48} & \heatmap{23.66} & \heatmap{16.83} & \heatmap{21.88} \\
  & & $\mathcal{A}_2$ & \heatmap{89.30} & \heatmap{57.66} & \heatmap{44.10} & \heatmap{49.61} & \heatmap{49.31} & \heatmap{21.97} & \heatmap{53.06} & \heatmap{41.23} & \heatmap{27.21} & \heatmap{19.92} & \heatmap{15.54} & \heatmap{10.62} & \heatmap{25.65} \\
  & & $\mathcal{A}_3$ & \heatmap{90.62} & \heatmap{63.35} & \heatmap{38.24} & \heatmap{48.28} & \heatmap{50.50} & \heatmap{21.86} & \heatmap{58.98} & \heatmap{35.99} & \heatmap{25.16} & \heatmap{20.13} & \heatmap{13.67} & \heatmap{10.19} & \heatmap{23.89} \\

  & &  $\mathcal{A}dj (Vot_{el})$ & \heatmap{90.34} & \heatmap{60.38} & \heatmap{38.66} & \heatmap{61.11} & \heatmap{58.62} & \heatmap{35.88} & \heatmap{55.95} & \heatmap{36.53} & \heatmap{25.00} & \heatmap{32.86} & \heatmap{21.36} & \heatmap{15.63} & \heatmap{23.58} \\
  & & $\mathcal{A}dj(V_{\approx})$ & \heatmap{85.08} & \heatmap{63.30} & \heatmap{32.27} & \heatmap{56.82} & \heatmap{52.31} & \heatmap{31.56} & \heatmap{58.85} & \heatmap{30.50} & \heatmap{22.91} & \heatmap{29.13} & \heatmap{22.48} & \heatmap{12.78} & \heatmap{21.66} \\
  
  & & $\mathcal{A}dj(V_{=})$ & \heatmap{51.65} & \heatmap{41.31} & \heatmap{25.05} & \heatmap{36.14} & \heatmap{33.10} & \heatmap{20.78} & \heatmap{38.71} & \heatmap{23.72} & \heatmap{19.53} & \heatmap{19.58} & \heatmap{16.95} & \heatmap{10.57} & \heatmap{18.46} \\
  
  & & $\mathcal{A}dj(\mathcal{A}_1)$ & \heatmap{90.85} & \heatmap{60.04} & \heatmap{34.75} & \heatmap{76.28} & \heatmap{60.71} & \heatmap{46.74} & \heatmap{55.77} & \heatmap{32.77} & \heatmap{22.59} & \heatmap{43.32} & \heatmap{27.08} & \heatmap{18.10} & \heatmap{21.37} \\
\cmidrule{1-1}\cmidrule{3-16}
\multirow{4}{*}{restaurant}
  & & $\mathcal{A}_1$ & \heatmap{90.45} & \heatmap{57.85} & \heatmap{46.60} & \heatmap{67.11} & \heatmap{78.10} & \heatmap{58.95} & \heatmap{54.22} & \heatmap{44.18} & \heatmap{34.11} & \heatmap{54.98} & \heatmap{37.52} & \heatmap{27.80} & \heatmap{32.59} \\
  & & $\mathcal{A}_2$ & \heatmap{87.93} & \heatmap{60.00} & \heatmap{49.27} & \heatmap{54.07} & \heatmap{68.22} & \heatmap{49.11} & \heatmap{56.03} & \heatmap{46.46} & \heatmap{37.32} & \heatmap{44.55} & \heatmap{36.23} & \heatmap{27.91} & \heatmap{35.69} \\
  & & $\mathcal{A}_3$ & \heatmap{90.26} & \heatmap{64.57} & \heatmap{50.73} & \heatmap{51.98} & \heatmap{67.26} & \heatmap{43.78} & \heatmap{60.42} & \heatmap{46.77} & \heatmap{36.58} & \heatmap{39.51} & \heatmap{28.45} & \heatmap{22.93} & \heatmap{34.34} \\

  &  &  $\mathcal{A}dj (Vot_{el})$ & \heatmap{89.67} & \heatmap{61.23} & \heatmap{49.27} & \heatmap{65.54} & \heatmap{77.17} & \heatmap{57.94} & \heatmap{57.30} & \heatmap{46.83} & \heatmap{37.32} & \heatmap{53.47} & \heatmap{39.53} & \heatmap{30.89} & \heatmap{35.68} \\
  & & $\mathcal{A}dj(V_{\approx})$ &   \heatmap{83.36} & \heatmap{63.62} & \heatmap{43.92} & \heatmap{62.64} & \heatmap{70.02} & \heatmap{55.62} & \heatmap{59.93} & \heatmap{41.81} & \heatmap{35.88} & \heatmap{52.36} & \heatmap{41.07} & \heatmap{29.49} & \heatmap{34.21} \\

  & & $\mathcal{A}dj(V_{=})$ & \heatmap{61.75} & \heatmap{49.49} & \heatmap{40.00} & \heatmap{51.21} & \heatmap{52.18} & \heatmap{46.90} & \heatmap{46.95} & \heatmap{38.26} & \heatmap{33.02} & \heatmap{44.24} & \heatmap{38.24} & \heatmap{29.70} & \heatmap{31.77} \\
  
  & & $\mathcal{A}dj(\mathcal{A}_1)$ & \heatmap{90.08} & \heatmap{64.18} & \heatmap{47.57} & \heatmap{79.63} & \heatmap{84.32} & \heatmap{71.48} & \heatmap{59.87} & \heatmap{44.72} & \heatmap{37.71} & \heatmap{66.00} & \heatmap{49.19} & \heatmap{37.46} & \heatmap{35.78} \\

\midrule

\multirow{4}{*}{laptop}
  & \multirow{13}{*}{\rotatebox{90}{med}} & $\mathcal{A}_1$ & \heatmap{89.86} & \heatmap{68.20} & \heatmap{55.29} & \heatmap{62.96} & \heatmap{53.05} & \heatmap{36.30} & \heatmap{62.67} & \heatmap{52.17} & \heatmap{39.57} & \heatmap{33.35} & \heatmap{25.95} & \heatmap{22.61} & \heatmap{37.46} \\
  & & $\mathcal{A}_2$ & \heatmap{90.37} & \heatmap{61.84} & \heatmap{43.96} & \heatmap{56.42} & \heatmap{56.75} & \heatmap{33.15} & \heatmap{57.22} & \heatmap{42.07} & \heatmap{29.04} & \heatmap{30.59} & \heatmap{20.72} & \heatmap{15.40} & \heatmap{27.66} \\
  & & $\mathcal{A}_3$ & \heatmap{89.30} & \heatmap{57.66} & \heatmap{44.10} & \heatmap{49.61} & \heatmap{49.31} & \heatmap{21.97} & \heatmap{53.06} & \heatmap{41.23} & \heatmap{27.21} & \heatmap{19.92} & \heatmap{15.54} & \heatmap{10.62} & \heatmap{25.65} \\

&  &  $\mathcal{A}dj (Vot_{el})$ & \heatmap{90.17} & \heatmap{66.60} & \heatmap{50.05} & \heatmap{61.46} & \heatmap{53.97} & \heatmap{34.87} & \heatmap{61.89} & \heatmap{47.46} & \heatmap{35.71} & \heatmap{32.28} & \heatmap{24.09} & \heatmap{18.71} & \heatmap{33.79} \\
  & & $\mathcal{A}dj(V_{\approx})$& \heatmap{86.19} & \heatmap{65.47} & \heatmap{39.38} & \heatmap{57.09} & \heatmap{50.11} & \heatmap{31.39} & \heatmap{61.07} & \heatmap{37.62} & \heatmap{29.55} & \heatmap{29.57} & \heatmap{23.28} & \heatmap{15.89} & \heatmap{28.04} \\

   & & $\mathcal{A}dj(V_{=})$& \heatmap{54.41} & \heatmap{44.31} & \heatmap{30.08} & \heatmap{38.64} & \heatmap{35.85} & \heatmap{24.22} & \heatmap{41.90} & \heatmap{28.90} & \heatmap{24.34} & \heatmap{22.88} & \heatmap{20.60} & \heatmap{14.42} & \heatmap{23.42} \\
  
  & & $\mathcal{A}dj(\mathcal{A}_1)$ & \heatmap{90.40} & \heatmap{73.53} & \heatmap{59.59} & \heatmap{70.51} & \heatmap{54.27} & \heatmap{41.23} & \heatmap{67.71} & \heatmap{56.60} & \heatmap{45.80} & \heatmap{37.90} & \heatmap{30.68} & \heatmap{24.84} & \heatmap{43.53} \\
\cmidrule{1-1}\cmidrule{3-16}
\multirow{4}{*}{restaurant}
  & & $\mathcal{A}_1$ & \heatmap{89.94} & \heatmap{69.39} & \heatmap{64.13} & \heatmap{65.41} & \heatmap{72.28} & \heatmap{54.91} & \heatmap{63.90} & \heatmap{60.16} & \heatmap{50.09} & \heatmap{49.01} & \heatmap{39.26} & \heatmap{37.66} & \heatmap{47.46} \\
  & & $\mathcal{A}_2$ & \heatmap{91.50} & \heatmap{63.30} & \heatmap{54.71} & \heatmap{52.97} & \heatmap{63.72} & \heatmap{46.25} & \heatmap{58.98} & \heatmap{52.52} & \heatmap{39.54} & \heatmap{42.48} & \heatmap{28.08} & \heatmap{25.09} & \heatmap{38.32} \\
  & & $\mathcal{A}_3$ & \heatmap{87.93} & \heatmap{60.00} & \heatmap{49.27} & \heatmap{54.07} & \heatmap{68.22} & \heatmap{49.11} & \heatmap{56.03} & \heatmap{46.46} & \heatmap{37.32} & \heatmap{44.55} & \heatmap{36.23} & \heatmap{27.91} & \heatmap{35.69} \\

  &   &  $\mathcal{A}dj (Vot_{el})$ & \heatmap{91.08} & \heatmap{69.29} & \heatmap{60.44} & \heatmap{65.23} & \heatmap{73.31} & \heatmap{56.22} & \heatmap{64.37} & \heatmap{57.55} & \heatmap{47.90} & \heatmap{51.58} & \heatmap{39.25} & \heatmap{36.29} & \heatmap{45.87} \\
  & & $\mathcal{A}dj(V_{\approx})$ & \heatmap{87.88} & \heatmap{67.69} & \heatmap{51.52} & \heatmap{59.43} & \heatmap{66.00} & \heatmap{49.67} & \heatmap{63.96} & \heatmap{49.90} & \heatmap{42.45} & \heatmap{46.14} & \heatmap{37.44} & \heatmap{29.14} & \heatmap{41.10} \\

  & & $\mathcal{A}dj(V_{=})$& \heatmap{63.03} & \heatmap{52.60} & \heatmap{46.53} & \heatmap{47.01} & \heatmap{50.73} & \heatmap{42.66} & \heatmap{49.71} & \heatmap{45.48} & \heatmap{39.41} & \heatmap{39.74} & \heatmap{36.29} & \heatmap{31.48} & \heatmap{38.47} \\
  
  & &  $\mathcal{A}dj(\mathcal{A}_1)$ & \heatmap{91.01} & \heatmap{72.74} & \heatmap{64.60} & \heatmap{82.96} & \heatmap{82.43} & \heatmap{71.75} & \heatmap{68.13} & \heatmap{61.17} & \heatmap{52.86} & \heatmap{66.40} & \heatmap{55.78} & \heatmap{48.85} & \heatmap{50.06} \\
\bottomrule
\end{tabular}}

\end{table}

\newpage
\section{Adjudication Mechanism: Full Breakdown}\label{sec:mechfull}
Per-cell decomposition of each emitted tuple into copy (\emph{match}), recombination, and generation (\emph{unique}), with the correct count and the derived generation share and synthesis-correct rate.
\paragraph{Full mechanism breakdown}\ Per-cell copy/recomb/generate counts with correct totals and derived \%gen and \%correct-from-synthesis, for the adjudicator and element-wise voting.
\begin{table}[!ht]
\scriptsize
\centering
\caption{Tuple output composition and correctness for ASTE (3-element) and ACOS (4-element). For each method, the upper row gives Total~(Correct) and the lower italic row gives that category's share of the method's overall output. Exact and Flex cannot produce recombined or unique tuples by construction; these cells are marked ``--''.}
\label{tab:tuple_composition_side_by_side}
\renewcommand{\arraystretch}{1.15}
\resizebox{\textwidth}{!}{%
\begin{tabular}{llcccr@{\hspace{1.5em}}cccr}
\toprule
& & \multicolumn{4}{c}{\textbf{ASTE (3-element)}} & \multicolumn{4}{c}{\textbf{ACOS (4-element)}} \\
\cmidrule(lr){3-6} \cmidrule(lr){7-10}
\textbf{Scale} & \textbf{Method} & \textbf{copy} & \textbf{recomb} & \textbf{gen} & \textbf{Total} & \textbf{copy} & \textbf{recomb} & \textbf{gen} & \textbf{Total} \\
\midrule
\multirow{8}{*}{\textbf{Med}} & \multirow{2}{*}{Exact} & 2527 (1590) & \multirow{2}{*}{--} & \multirow{2}{*}{--} & \multirow{2}{*}{\textbf{2527}} & 792 (272) & \multirow{2}{*}{--} & \multirow{2}{*}{--} & \multirow{2}{*}{\textbf{792}} \\
 &  & \textit{100.0\%} &  &  &  & \textit{100.0\%} &  &  &  \\
 & \multirow{2}{*}{Flex} & 2713 (1588) & \multirow{2}{*}{--} & \multirow{2}{*}{--} & \multirow{2}{*}{\textbf{2713}} & 1728 (325) & \multirow{2}{*}{--} & \multirow{2}{*}{--} & \multirow{2}{*}{\textbf{1728}} \\
 &  & \textit{100.0\%} &  &  &  & \textit{100.0\%} &  &  &  \\
 & \multirow{2}{*}{Elem} & 2868 (1639) & 19 (0) & \multirow{2}{*}{--} & \multirow{2}{*}{\textbf{2887}} & 1843 (387) & 198 (7) & \multirow{2}{*}{--} & \multirow{2}{*}{\textbf{2041}} \\
 &  & \textit{99.3\%} & \textit{0.7\%} &  &  & \textit{90.3\%} & \textit{9.7\%} &  &  \\
 & \multirow{2}{*}{LLM} & 2643 (1651) & 17 (2) & 98 (9) & \multirow{2}{*}{\textbf{2758}} & 1133 (451) & 291 (50) & 711 (84) & \multirow{2}{*}{\textbf{2135}} \\
 &  & \textit{95.8\%} & \textit{0.6\%} & \textit{3.6\%} &  & \textit{53.1\%} & \textit{13.6\%} & \textit{33.3\%} &  \\
\cmidrule{1-10}
\multirow{8}{*}{\textbf{Small}} & \multirow{2}{*}{Exact} & 2392 (1494) & \multirow{2}{*}{--} & \multirow{2}{*}{--} & \multirow{2}{*}{\textbf{2392}} & 748 (225) & \multirow{2}{*}{--} & \multirow{2}{*}{--} & \multirow{2}{*}{\textbf{748}} \\
 &  & \textit{100.0\%} &  &  &  & \textit{100.0\%} &  &  &  \\
 & \multirow{2}{*}{Flex} & 2620 (1505) & \multirow{2}{*}{--} & \multirow{2}{*}{--} & \multirow{2}{*}{\textbf{2620}} & 1613 (285) & \multirow{2}{*}{--} & \multirow{2}{*}{--} & \multirow{2}{*}{\textbf{1613}} \\
 &  & \textit{100.0\%} &  &  &  & \textit{100.0\%} &  &  &  \\
 & \multirow{2}{*}{Elem} & 2805 (1555) & 23 (0) & \multirow{2}{*}{--} & \multirow{2}{*}{\textbf{2828}} & 1761 (304) & 169 (4) & \multirow{2}{*}{--} & \multirow{2}{*}{\textbf{1930}} \\
 &  & \textit{99.2\%} & \textit{0.8\%} &  &  & \textit{91.2\%} & \textit{8.8\%} &  &  \\
 & \multirow{2}{*}{LLM} & 2834 (1522) & 36 (0) & 12 (0) & \multirow{2}{*}{\textbf{2882}} & 1540 (360) & 226 (11) & 269 (20) & \multirow{2}{*}{\textbf{2035}} \\
 &  & \textit{98.3\%} & \textit{1.2\%} & \textit{0.4\%} &  & \textit{75.7\%} & \textit{11.1\%} & \textit{13.2\%} &  \\
\cmidrule{1-10}
\multirow{8}{*}{\textbf{Mini}} & \multirow{2}{*}{Exact} & 1976 (1354) & \multirow{2}{*}{--} & \multirow{2}{*}{--} & \multirow{2}{*}{\textbf{1976}} & 469 (220) & \multirow{2}{*}{--} & \multirow{2}{*}{--} & \multirow{2}{*}{\textbf{469}} \\
 &  & \textit{100.0\%} &  &  &  & \textit{100.0\%} &  &  &  \\
 & \multirow{2}{*}{Flex} & 2323 (1366) & \multirow{2}{*}{--} & \multirow{2}{*}{--} & \multirow{2}{*}{\textbf{2323}} & 1437 (290) & \multirow{2}{*}{--} & \multirow{2}{*}{--} & \multirow{2}{*}{\textbf{1437}} \\
 &  & \textit{100.0\%} &  &  &  & \textit{100.0\%} &  &  &  \\
 & \multirow{2}{*}{Elem} & 2711 (1515) & 42 (1) & \multirow{2}{*}{--} & \multirow{2}{*}{\textbf{2753}} & 1806 (390) & 159 (11) & \multirow{2}{*}{--} & \multirow{2}{*}{\textbf{1965}} \\
 &  & \textit{98.5\%} & \textit{1.5\%} &  &  & \textit{91.9\%} & \textit{8.1\%} &  &  \\
 & \multirow{2}{*}{LLM} & 2492 (1553) & 60 (12) & 98 (17) & \multirow{2}{*}{\textbf{2650}} & 1487 (387) & 116 (23) & 365 (53) & \multirow{2}{*}{\textbf{1968}} \\
 &  & \textit{94.0\%} & \textit{2.3\%} & \textit{3.7\%} &  & \textit{75.6\%} & \textit{5.9\%} & \textit{18.5\%} &  \\
\bottomrule
\end{tabular}
}
\end{table}

\newpage
\section{Test For Statistical Significance}\label{sec:significance}
Tables~\ref{tab:aste_results_updated} and~\ref{tab:acos_results_updated} report paired-bootstrap comparisons of each aggregation method ($\mathcal{A}dj(Vot_{\approx})$, $\mathcal{A}dj(Vot_{=})$, $\mathcal{A}dj(Vot_{\mathrm{el}})$, and the LLM adjudicator $\mathcal{A}dj(\mathcal{A}_1)$) against each individual annotator $\mathcal{A}_1$--$\mathcal{A}_3$. Each cell gives the mean F1 difference (aggregated $-$ individual, scaled to 0--100) with its 95\% confidence interval and $p$-value; green shading marks a significant positive gain ($p<0.05$). Green highlights that the adjudication performance is better than at least 2 of the individual annotators; yellow highlights mark that the adjudication performance is significantly better than all individual annotators.

\begin{table*}[htbp]
\centering
\caption{Statistical Significance Test Results - ASTE (scaled to 0--100)}
\label{tab:aste_results_updated}
\resizebox{0.96\textwidth}{!}{%
\begin{tabular}{@{}lllccc@{}}
\toprule
\textbf{Model} & \textbf{Dataset} & \textbf{Method} & \textbf{A1} & \textbf{A2} & \textbf{A3} \\
\midrule
\multirow{16}{*}{MINI} & \multirow{4}{*}{LAP14} & $\mathcal{A}dj(V_{\approx})$ & -2.90 [-6.26, +0.32], p=0.0829 & \cellcolor{green!25}+3.97 [+0.86, +7.14], p=0.0135 & +16.49 [+12.78, +20.21], p=0.0000 \\
 &  & $\mathcal{A}dj(V_{=})$ & +0.73 [-2.73, +3.99], p=0.6758 & \cellcolor{green!25}+7.60 [+4.33, +10.86], p=0.0000 & +20.12 [+16.36, +23.80], p=0.0000 \\
 &  & $\mathcal{A}dj(V_{\mathrm{el}})$ & -1.72 [-3.86, +0.39], p=0.1159 & \cellcolor{green!25}+5.15 [+1.48, +8.89], p=0.0075 & +17.67 [+13.91, +21.53], p=0.0000 \\
 &  & $\mathcal{A}dj(\mathcal{A}_1)$ & \cellcolor{yellow!40}+2.87 [+0.34, +5.48], p=0.0297 & \cellcolor{green!25}+9.75 [+5.71, +14.02], p=0.0000 & +22.26 [+17.79, +26.80], p=0.0000 \\
\cmidrule(l){2-6}
 & \multirow{4}{*}{RES14} & $\mathcal{A}dj(V_{\approx})$ & -2.84 [-5.35, -0.43], p=0.0229 & \cellcolor{green!25}+3.68 [+0.96, +6.30], p=0.0063 & +22.81 [+19.69, +26.10], p=0.0000 \\
 &  & $\mathcal{A}dj(V_{=})$ & +0.61 [-1.73, +2.85], p=0.6109 & \cellcolor{green!25}+7.13 [+4.45, +9.69], p=0.0000 & +26.26 [+23.08, +29.58], p=0.0000 \\
 &  & $\mathcal{A}dj(V_{\mathrm{el}})$ & -0.32 [-2.18, +1.62], p=0.7340 & \cellcolor{green!25}+6.20 [+3.35, +9.11], p=0.0001 & +25.34 [+21.98, +28.83], p=0.0000 \\
 &  & $\mathcal{A}dj(\mathcal{A}_1)$ & \cellcolor{yellow!40}+3.51 [+1.44, +5.68], p=0.0017 & \cellcolor{green!25}+10.03 [+6.40, +13.59], p=0.0001 & +29.17 [+25.31, +33.06], p=0.0000 \\
\cmidrule(l){2-6}
 & \multirow{4}{*}{RES15} & $\mathcal{A}dj(V_{\approx})$ & -2.38 [-5.45, +0.57], p=0.1235 & \cellcolor{green!25}+4.94 [+1.98, +7.93], p=0.0014 & +17.85 [+13.97, +21.89], p=0.0000 \\
 &  & $\mathcal{A}dj(V_{=})$ & +1.30 [-2.04, +4.51], p=0.4338 & \cellcolor{green!25}+8.61 [+5.56, +11.64], p=0.0000 & +21.52 [+17.56, +25.47], p=0.0000 \\
 &  & $\mathcal{A}dj(V_{\mathrm{el}})$ & -2.03 [-4.66, +0.46], p=0.1171 & \cellcolor{green!25}+5.29 [+2.14, +8.59], p=0.0010 & +18.20 [+14.22, +22.36], p=0.0000 \\
 &  & $\mathcal{A}dj(\mathcal{A}_1)$ & \cellcolor{yellow!40}+3.28 [+0.32, +6.22], p=0.0315 & \cellcolor{green!25}+10.60 [+6.64, +14.58], p=0.0000 & +23.51 [+19.22, +27.91], p=0.0000 \\
\cmidrule(l){2-6}
 & \multirow{4}{*}{RES16} & $\mathcal{A}dj(V_{\approx})$ & +0.37 [-2.03, +2.81], p=0.7596 & \cellcolor{green!25}+7.74 [+3.65, +11.80], p=0.0005 & +21.51 [+17.75, +25.36], p=0.0000 \\
 &  & $\mathcal{A}dj(V_{=})$ & \cellcolor{yellow!40}+4.43 [+2.03, +6.94], p=0.0005 & \cellcolor{green!25}+11.80 [+7.86, +15.84], p=0.0000 & +25.57 [+21.93, +29.34], p=0.0000 \\
 &  & $\mathcal{A}dj(V_{\mathrm{el}})$ & +0.48 [-1.28, +2.35], p=0.6007 & \cellcolor{green!25}+7.85 [+3.44, +12.40], p=0.0012 & +21.62 [+17.78, +25.54], p=0.0000 \\
 &  & $\mathcal{A}dj(\mathcal{A}_1)$ & +1.41 [-0.67, +3.48], p=0.1864 & \cellcolor{green!25}+8.78 [+4.19, +13.53], p=0.0006 & +22.55 [+18.32, +26.76], p=0.0000 \\
\midrule
\multirow{16}{*}{MSMALL} & \multirow{4}{*}{LAP14} & $\mathcal{A}dj(V_{\approx})$ & -2.10 [-4.70, +0.38], p=0.1059 & +1.75 [-0.71, +4.26], p=0.1667 & +9.88 [+6.89, +12.90], p=0.0000 \\
 &  & $\mathcal{A}dj(V_{=})$ & +0.11 [-2.64, +2.71], p=0.9388 & \cellcolor{green!25}+3.95 [+1.52, +6.39], p=0.0020 & +12.09 [+9.07, +15.18], p=0.0000 \\
 &  & $\mathcal{A}dj(V_{\mathrm{el}})$ & -2.87 [-5.01, -0.87], p=0.0081 & +0.98 [-1.65, +3.70], p=0.4734 & +9.11 [+6.01, +12.21], p=0.0000 \\
 &  & $\mathcal{A}dj(\mathcal{A}_1)$ & -4.64 [-7.11, -2.30], p=0.0001 & -0.80 [-3.19, +1.67], p=0.5225 & +7.34 [+4.30, +10.43], p=0.0000 \\
\cmidrule(l){2-6}
 & \multirow{4}{*}{RES14} & $\mathcal{A}dj(V_{\approx})$ & +0.42 [-1.83, +2.78], p=0.7175 & \cellcolor{green!25}+4.13 [+1.86, +6.33], p=0.0005 & +6.24 [+3.91, +8.57], p=0.0000 \\
 &  & $\mathcal{A}dj(V_{=})$ & \cellcolor{yellow!40}+2.61 [+0.35, +4.98], p=0.0302 & \cellcolor{green!25}+6.32 [+4.10, +8.51], p=0.0000 & +8.42 [+6.05, +10.77], p=0.0000 \\
 &  & $\mathcal{A}dj(V_{\mathrm{el}})$ & +0.37 [-1.76, +2.59], p=0.7342 & \cellcolor{green!25}+4.08 [+1.96, +6.21], p=0.0006 & +6.19 [+3.80, +8.59], p=0.0000 \\
 &  & $\mathcal{A}dj(\mathcal{A}_1)$ & -1.77 [-3.87, +0.41], p=0.1059 & +1.94 [-0.53, +4.33], p=0.1196 & +4.05 [+1.45, +6.64], p=0.0022 \\
\cmidrule(l){2-6}
 & \multirow{4}{*}{RES15} & $\mathcal{A}dj(V_{\approx})$ & -3.10 [-5.74, -0.48], p=0.0219 & \cellcolor{green!25}+3.24 [+0.24, +6.15], p=0.0337 & +6.66 [+3.67, +9.69], p=0.0000 \\
 &  & $\mathcal{A}dj(V_{=})$ & -0.59 [-3.06, +1.80], p=0.6348 & \cellcolor{green!25}+5.74 [+2.83, +8.53], p=0.0001 & +9.16 [+6.00, +12.30], p=0.0000 \\
 &  & $\mathcal{A}dj(V_{\mathrm{el}})$ & -2.67 [-4.86, -0.61], p=0.0139 & \cellcolor{green!25}+3.66 [+0.79, +6.49], p=0.0138 & +7.09 [+3.74, +10.47], p=0.0000 \\
 &  & $\mathcal{A}dj(\mathcal{A}_1)$ & -3.75 [-6.33, -1.22], p=0.0045 & +2.58 [-0.26, +5.35], p=0.0721 & +6.01 [+2.73, +9.32], p=0.0003 \\
\cmidrule(l){2-6}
 & \multirow{4}{*}{RES16} & $\mathcal{A}dj(V_{\approx})$ & +0.52 [-2.71, +3.66], p=0.7543 & \cellcolor{green!25}+6.53 [+4.06, +9.21], p=0.0000 & +3.07 [-0.59, +6.52], p=0.0925 \\
 &  & $\mathcal{A}dj(V_{=})$ & +2.49 [-0.76, +5.71], p=0.1310 & \cellcolor{green!25}+8.51 [+6.02, +11.20], p=0.0000 & +5.04 [+0.96, +8.74], p=0.0117 \\
 &  & $\mathcal{A}dj(V_{\mathrm{el}})$ & -0.38 [-3.38, +2.56], p=0.7949 & \cellcolor{green!25}+5.63 [+3.24, +8.25], p=0.0000 & +2.16 [-1.78, +5.86], p=0.2600 \\
 &  & $\mathcal{A}dj(\mathcal{A}_1)$ & -2.29 [-5.35, +0.67], p=0.1347 & \cellcolor{green!25}+3.72 [+1.16, +6.31], p=0.0061 & +0.25 [-3.33, +3.66], p=0.8844 \\
\midrule
\multirow{16}{*}{MED} & \multirow{4}{*}{LAP14} & $\mathcal{A}dj(V_{\approx})$ & -0.81 [-3.36, +1.67], p=0.5269 & +1.35 [-1.60, +4.32], p=0.3640 & +3.76 [+1.17, +6.40], p=0.0040 \\
 &  & $\mathcal{A}dj(V_{=})$ & +2.12 [-0.33, +4.60], p=0.0935 & \cellcolor{green!25}+4.28 [+1.34, +7.31], p=0.0046 & +6.69 [+4.11, +9.38], p=0.0000 \\
 &  & $\mathcal{A}dj(V_{\mathrm{el}})$ & +0.71 [-1.33, +2.78], p=0.4969 & +2.87 [-0.22, +6.13], p=0.0741 & +5.28 [+2.45, +8.17], p=0.0000 \\
 &  & $\mathcal{A}dj(\mathcal{A}_1)$ & +1.16 [-1.51, +3.83], p=0.3962 & +3.33 [-0.00, +6.73], p=0.0530 & +5.73 [+2.76, +8.80], p=0.0006 \\
\cmidrule(l){2-6}
 & \multirow{4}{*}{RES14} & $\mathcal{A}dj(V_{\approx})$ & -1.97 [-3.96, +0.02], p=0.0520 & \cellcolor{green!25}+4.24 [+1.38, +6.96], p=0.0036 & +6.63 [+4.54, +8.71], p=0.0000 \\
 &  & $\mathcal{A}dj(V_{=})$ & +0.32 [-1.72, +2.33], p=0.7608 & \cellcolor{green!25}+6.53 [+3.67, +9.20], p=0.0001 & +8.92 [+6.68, +11.17], p=0.0000 \\
 &  & $\mathcal{A}dj(V_{\mathrm{el}})$ & -2.10 [-3.99, -0.21], p=0.0272 & \cellcolor{green!25}+4.11 [+1.27, +6.78], p=0.0043 & +6.50 [+4.16, +8.86], p=0.0000 \\
 &  & $\mathcal{A}dj(\mathcal{A}_1)$ & +0.60 [-1.17, +2.39], p=0.5121 & \cellcolor{green!25}+6.82 [+3.77, +9.78], p=0.0000 & +9.21 [+6.53, +11.90], p=0.0000 \\
\cmidrule(l){2-6}
 & \multirow{4}{*}{RES15} & $\mathcal{A}dj(V_{\approx})$ & -0.69 [-3.19, +1.66], p=0.5679 & +1.03 [-1.38, +3.44], p=0.4078 & +7.33 [+4.38, +10.27], p=0.0000 \\
 &  & $\mathcal{A}dj(V_{=})$ & +2.06 [-0.49, +4.62], p=0.1134 & \cellcolor{green!25}+3.79 [+1.35, +6.19], p=0.0024 & +10.08 [+7.17, +12.95], p=0.0000 \\
 &  & $\mathcal{A}dj(V_{\mathrm{el}})$ & -0.57 [-2.88, +1.75], p=0.6209 & +1.15 [-1.48, +3.82], p=0.3972 & +7.45 [+4.58, +10.38], p=0.0000 \\
 &  & $\mathcal{A}dj(\mathcal{A}_1)$ & +0.53 [-2.03, +3.17], p=0.6941 & +2.25 [-0.65, +5.16], p=0.1233 & +8.55 [+5.23, +11.94], p=0.0000 \\
\cmidrule(l){2-6}
 & \multirow{4}{*}{RES16} & $\mathcal{A}dj(V_{\approx})$ & -1.54 [-4.66, +1.53], p=0.3295 & +0.80 [-1.60, +3.13], p=0.5066 & +4.86 [+2.24, +7.64], p=0.0004 \\
 &  & $\mathcal{A}dj(V_{=})$ & -0.74 [-3.81, +2.26], p=0.6355 & +1.60 [-0.75, +3.86], p=0.1677 & +5.66 [+3.08, +8.39], p=0.0000 \\
 &  & $\mathcal{A}dj(V_{\mathrm{el}})$ & -2.94 [-5.86, -0.14], p=0.0445 & -0.60 [-2.93, +1.67], p=0.5951 & +3.46 [+0.92, +6.11], p=0.0079 \\
 &  & $\mathcal{A}dj(\mathcal{A}_1)$ & +1.77 [-0.91, +4.64], p=0.2118 & \cellcolor{green!25}+4.11 [+1.17, +7.17], p=0.0085 & +8.17 [+5.02, +11.51], p=0.0000 \\
\bottomrule
\end{tabular}%
}
\end{table*}

\begin{table*}[htbp]
\centering
\caption{Statistical Significance Test Results - ACOS (scaled to 0--100)}
\label{tab:acos_results_updated}
\resizebox{0.9\textwidth}{!}{%
\begin{tabular}{@{}lllccc@{}}
\toprule
\textbf{Model} & \textbf{Domain} & \textbf{Method} & \textbf{A1} & \textbf{A2} & \textbf{A3} \\
\midrule
\multirow{8}{*}{MINI} & \multirow{4}{*}{Laptop} & $\mathcal{A}dj(V_{\approx})$ & -3.69 [-5.39, -2.06], p=0.0001 & -2.91 [-4.65, -1.24], p=0.0009 & +7.63 [+5.94, +9.36], p=0.0000 \\
 &  & $\mathcal{A}dj(V_{=})$ & -2.57 [-4.63, -0.57], p=0.0118 & -1.79 [-3.61, -0.04], p=0.0518 & +8.75 [+6.64, +10.98], p=0.0000 \\
 &  & $\mathcal{A}dj(V_{\mathrm{el}})$ & +0.27 [-1.00, +1.53], p=0.6672 & +1.06 [-0.96, +3.09], p=0.3142 & +11.59 [+9.61, +13.64], p=0.0000 \\
 &  & $\mathcal{A}dj(\mathcal{A}_1)$ & \cellcolor{yellow!40}+3.27 [+1.89, +4.72], p=0.0000 & \cellcolor{green!25}+4.05 [+1.81, +6.33], p=0.0008 & +14.59 [+12.23, +16.94], p=0.0000 \\
\cmidrule(l){2-6}
 & \multirow{4}{*}{Restaurant} & $\mathcal{A}dj(V_{\approx})$ & -4.13 [-6.22, -2.13], p=0.0002 & \cellcolor{green!25}+4.91 [+1.81, +7.94], p=0.0011 & +21.11 [+17.89, +24.36], p=0.0000 \\
 &  & $\mathcal{A}dj(V_{=})$ & -4.78 [-7.65, -1.97], p=0.0008 & \cellcolor{green!25}+4.25 [+1.62, +6.80], p=0.0017 & +20.45 [+16.82, +24.14], p=0.0000 \\
 &  & $\mathcal{A}dj(V_{\mathrm{el}})$ & -1.53 [-3.46, +0.37], p=0.1200 & \cellcolor{green!25}+7.50 [+4.55, +10.41], p=0.0000 & +23.70 [+20.57, +26.94], p=0.0000 \\
 &  & $\mathcal{A}dj(\mathcal{A}_1)$ & +1.32 [-1.96, +4.62], p=0.4292 & \cellcolor{green!25}+10.35 [+6.42, +14.21], p=0.0000 & +26.55 [+23.13, +29.92], p=0.0000 \\
\midrule
\multirow{8}{*}{MSMALL} & \multirow{4}{*}{Laptop} & $\mathcal{A}dj(V_{\approx})$ & -0.60 [-1.72, +0.50], p=0.2855 & +1.34 [-0.18, +2.93], p=0.0891 & +2.31 [+0.70, +3.96], p=0.0049 \\
 &  & $\mathcal{A}dj(V_{=})$ & -1.46 [-3.28, +0.29], p=0.1107 & +0.48 [-0.90, +1.87], p=0.4882 & +1.45 [+0.02, +2.94], p=0.0513 \\
 &  & $\mathcal{A}dj(V_{\mathrm{el}})$ & -0.83 [-2.00, +0.32], p=0.1640 & +1.12 [-0.28, +2.55], p=0.1208 & +2.08 [+0.54, +3.64], p=0.0077 \\
 &  & $\mathcal{A}dj(\mathcal{A}_1)$ & \cellcolor{yellow!40}+1.31 [+0.12, +2.50], p=0.0321 & \cellcolor{green!25}+3.26 [+1.61, +5.01], p=0.0002 & +4.22 [+2.45, +5.99], p=0.0000 \\
\cmidrule(l){2-6}
 & \multirow{4}{*}{Restaurant} & $\mathcal{A}dj(V_{\approx})$ & \cellcolor{yellow!40}+3.00 [+0.94, +5.13], p=0.0054 & +1.32 [-1.31, +4.07], p=0.3377 & +8.39 [+5.17, +11.69], p=0.0000 \\
 &  & $\mathcal{A}dj(V_{=})$ & \cellcolor{yellow!40}+3.98 [+1.47, +6.47], p=0.0017 & +2.30 [-0.50, +5.00], p=0.1032 & +9.38 [+6.19, +12.58], p=0.0000 \\
 &  & $\mathcal{A}dj(V_{\mathrm{el}})$ & \cellcolor{yellow!40}+3.00 [+1.01, +5.05], p=0.0030 & +1.32 [-1.28, +4.01], p=0.3250 & +8.40 [+5.14, +11.62], p=0.0000 \\
 &  & $\mathcal{A}dj(\mathcal{A}_1)$ & \cellcolor{yellow!40}+8.51 [+5.79, +11.39], p=0.0000 & \cellcolor{green!25}+6.84 [+4.16, +9.64], p=0.0000 & +13.91 [+10.79, +17.09], p=0.0000 \\
\midrule
\multirow{8}{*}{MED} & \multirow{4}{*}{Laptop} & $\mathcal{A}dj(V_{\approx})$ & -3.74 [-5.36, -2.14], p=0.0000 & \cellcolor{green!25}+2.23 [+0.57, +3.92], p=0.0093 & +4.27 [+2.47, +6.14], p=0.0000 \\
 &  & $\mathcal{A}dj(V_{=})$ & -3.61 [-5.66, -1.59], p=0.0009 & \cellcolor{green!25}+2.36 [+0.76, +4.03], p=0.0047 & +4.40 [+2.60, +6.27], p=0.0000 \\
 &  & $\mathcal{A}dj(V_{\mathrm{el}})$ & -2.77 [-4.30, -1.31], p=0.0001 & \cellcolor{green!25}+3.20 [+1.46, +5.02], p=0.0001 & +5.24 [+3.47, +7.08], p=0.0000 \\
 &  & $\mathcal{A}dj(\mathcal{A}_1)$ & \cellcolor{yellow!40}+2.99 [+0.63, +5.34], p=0.0139 & \cellcolor{green!25}+8.96 [+6.53, +11.29], p=0.0000 & +11.00 [+8.54, +13.42], p=0.0000 \\
\cmidrule(l){2-6}
 & \multirow{4}{*}{Restaurant} & $\mathcal{A}dj(V_{\approx})$ & -4.27 [-6.71, -1.87], p=0.0004 & \cellcolor{green!25}+6.55 [+4.00, +9.09], p=0.0000 & +1.00 [-2.55, +4.50], p=0.5819 \\
 &  & $\mathcal{A}dj(V_{=})$ & -0.79 [-3.60, +1.98], p=0.5701 & \cellcolor{green!25}+10.02 [+7.46, +12.52], p=0.0000 & +4.47 [+0.87, +7.98], p=0.0157 \\
 &  & $\mathcal{A}dj(V_{\mathrm{el}})$ & -0.76 [-2.52, +1.05], p=0.3925 & \cellcolor{green!25}+10.05 [+7.04, +13.02], p=0.0000 & +4.50 [+0.68, +8.32], p=0.0209 \\
 &  & $\mathcal{A}dj(\mathcal{A}_1)$ & \cellcolor{yellow!40}+11.91 [+8.53, +15.30], p=0.0000 & \cellcolor{green!25}+22.72 [+18.78, +26.48], p=0.0000 & +17.17 [+13.97, +20.46], p=0.0000 \\
\bottomrule
\end{tabular}%
}
\end{table*}

\newpage
\section{Recovered Pipeline Throughput}\label{sec:throughput}

\begin{table}[thbp]
\centering
\footnotesize
\caption{Per-sentence latency on a single A40, reconstructed from
output-file timestamps. Ranges span shot settings (per-group means and
medians); counts are timed files after outlier removal. The upper blocks
cover the annotators and adjudicators and
are the sole basis for the cost figures in Section~\ref{sec:setup}. The
final block reports scale-matched runs on models not used elsewhere in
this work; these are cited only to establish that ACOS is the slower
task, and contribute to no reported figure. DS = DeepSeek.}
\label{tab:throughput}
\begin{tabular}{llllrrr}
\toprule
Task & Pass & Model & Shots & Sent. & Mean (s) & Median (s) \\
\midrule
\multicolumn{7}{l}{\textit{Annotator passes (Table~\ref{tab:tdatasets} models)}} \\
ACOS & annot. & DS-R1-Distill-Qwen-32B    & 0/5/10/15 & 2{,}395 & 63.5--69.4 & 61.6--67.7 \\
ACOS & annot. & Qwen3-4B-Thinking         & 15        & 1{,}399 & 77.5       & 76.4 \\
\midrule
\multicolumn{7}{l}{\textit{Adjudication passes (Table~\ref{tab:tdatasets} models)}} \\
ASTE & adjud. & Qwen3-30B-A3B-Think.-FP8  & 0/5/10/15 & 2{,}544 & 28.9--58.7 & 23.9--44.6 \\
ASTE & adjud. & Qwen3-14B                 & 5         & 326     & 25.4       & 19.6 \\
ASTE & adjud. & Qwen3-4B-Thinking         & 5         & 1{,}461 & 45.9       & 39.2 \\
ACOS & adjud. & Qwen3-30B-A3B-Think.-FP8  & 5/10/15   & 1{,}840 & 24.9--27.0 & 20.6--25.0 \\
ACOS & adjud. & Qwen3-14B                 & 15        & 1{,}399 & 42.7       & 39.0 \\
ACOS & adjud. & Qwen3-4B-Thinking         & 5         & 1{,}370 & 50.7       & 48.2 \\
\midrule
\multicolumn{7}{l}{\textit{Reference runs, scale-matched (not used in this work; task-ordering evidence only)}} \\
ASTE & annot. & Qwen3-8B                  & 0/5/10/15 & 3{,}104 & 14.7--17.1 & 12.3--14.2 \\
ACOS & annot. & Qwen3-8B                  & 0/5/10/15 & 2{,}395 & 17.6--27.4 & 16.0--26.0 \\
ASTE & annot. & DS-R1-0528-Qwen3-8B       & 0/5/10/15 & 3{,}077 & 16.7--34.7 & 16.1--20.3 \\
ACOS & annot. & DS-R1-0528-Qwen3-8B       & 0/5/10/15 & 996     & 21.5--99.0 & 18.8--96.1 \\
\bottomrule
\end{tabular}
\end{table}

Per-sentence latency was derived from modification-timestamp gaps across 79 intact runs on a single NVIDIA A40 (48\,GB). Outliers ($>Q_3 + 1.5\times\mathrm{IQR}$; 1.5--9\% of gaps) were discarded as restart stalls. Timings are pooled strictly within (task, model, step, shots) groups, excluding small folders and sub-2\,s write-buffered runs,
yielding $\sim$22.3k timed annotations (Table~\ref{tab:throughput}), of which 12.7k cover the models used in this work. Paired same-model rows show ACOS consistently slower than ASTE at equal shot settings (e.g.\ Qwen3-8B: 14.7--17.1\,s vs.\ 17.6--27.4\,s), which is why the ACOS-only annotator timings yield a conservative cost bound. Few-shot prompting increases latency up to $2\times$ for large adjudicators
(Qwen3-30B on ASTE: 28.9\,s zero-shot vs.\ 58.7\,s 10-shot), with modest
impact on smaller models. Costs use July~2026 A40 rates
(\$0.29--\$0.44/h; \pounds1 $\approx$ \$1.34).
\end{document}